\documentclass[10pt,twocolumn,letterpaper]{article}

 \usepackage[pagenumbers]{cvpr} 

\definecolor{cvprblue}{rgb}{0.21,0.49,0.74}
\usepackage[pagebackref,breaklinks,colorlinks,allcolors=cvprblue]{hyperref}
\usepackage{caption}
\usepackage{pifont}
\usepackage{multirow}
\usepackage{colortbl}
\usepackage{tabularx}
\usepackage{array} 
\usepackage{graphicx}

\makeatletter
\def\blfootnote{\xdef\@thefnmark{*}\@footnotetext}
\makeatother

\definecolor{best}{rgb}{1, 0.5, 0}
\definecolor{second}{rgb}{1, 1, 0}

\definecolor{fig1green}{rgb}{0.153, 0.378, 0.227}
\definecolor{fig1pink}{rgb}{0.871, 0.706, 0.745}
\definecolor{fig2green}{rgb}{0.455, 0.886, 0.553}

\usepackage[capitalize]{cleveref}
\crefname{section}{Sec.}{Secs.}
\Crefname{section}{Section}{Sections}
\Crefname{table}{Table}{Tables}
\crefname{table}{Tab.}{Tabs.}

\def\paperID{3505} 
\def\confName{CVPR}
\def\confYear{2025}

\title{CoCoGaussian: Leveraging Circle of Confusion for \\Gaussian Splatting from Defocused Images\vspace{-2mm}}

\author{{\fontsize{11}{11}\selectfont Jungho Lee$^{1*}$ \quad Suhwan Cho$^1$ \quad Taeoh Kim$^2$ \quad Ho-Deok Jang$^2$ \quad Minhyeok Lee$^1$} \\ {\fontsize{11}{11}\selectfont Geonho Cha$^2$ \quad Dongyoon Wee$^2$ \quad Dogyoon Lee$^1$ \quad Sangyoun Lee$^1$}\vspace{2mm}\\
	{\fontsize{11}{11}\selectfont$^1$School of Electrical and Electronic Engineering, Yonsei University} \quad	{\fontsize{11}{11}\selectfont$^2$NAVER Cloud}\\
	{\fontsize{10}{10}\selectfont \tt \textbf{\href{https://Jho-Yonsei.github.io/CoCoGaussian/}{\texttt{https://Jho-Yonsei.github.io/CoCoGaussian}}}}
}

\begin{document}
\twocolumn[{
	\renewcommand\twocolumn[1][]{#1}
	\maketitle
	\begin{center}
		\centering
		\captionsetup{type=figure}
		\vspace{-8mm}
		\includegraphics[width=0.95\linewidth]{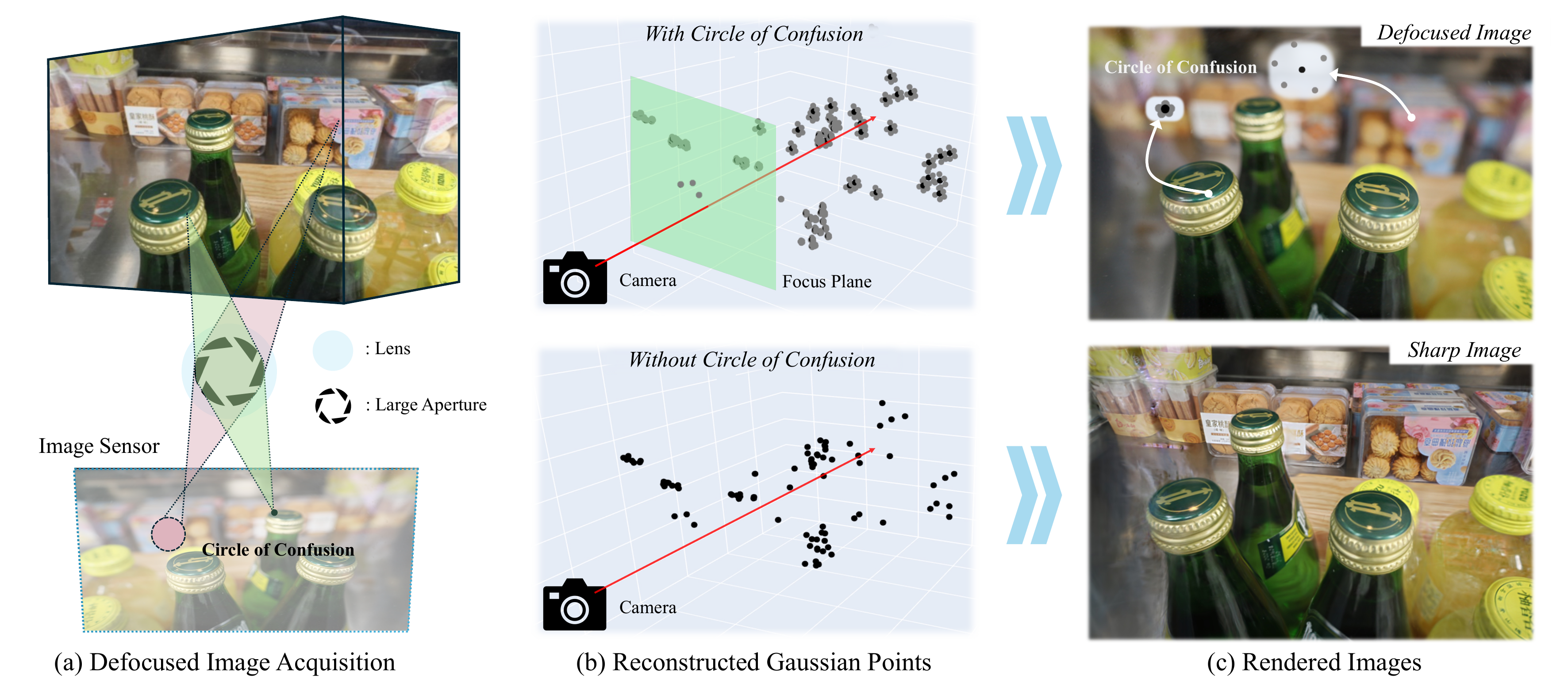}
		\vspace{-2mm}
		\caption{We propose CoCoGaussian, a novel framework for 3D scene reconstruction from defocused images. \textbf{(a)} With a large aperture, radiance from the focus plane appears as a small circle of confusion (\textbf{\textcolor{fig1green}{green}}) on the image sensor, while radiance from greater depths results in larger circles (\textbf{\textcolor{fig1pink}{pink}}). \textbf{(b)} We visualize reconstructed Gaussians of CoCoGaussian, where the black dots indicate Gaussian means, while the gray dots represent Gaussian means forming the circle of confusion, which decreases in size near the focus plane. \textbf{(c)} The defocused image focuses on shallow depths, with circle size increasing with depth, as shown in (b). CoCoGaussian allows customization of defocused images by adjusting the depth of field or focus plane, while sharp images can be obtained by rendering without the circle of confusion.}
		\label{fig:teaser}
	\end{center}
}]

\begin{abstract}
\blfootnote{This work was done during an internship at NAVER Cloud.} 3D Gaussian Splatting (3DGS) has attracted significant attention for its high-quality novel view rendering, inspiring research to address real-world challenges. While conventional methods depend on sharp images for accurate scene reconstruction, real-world scenarios are often affected by defocus blur due to finite depth of field, making it essential to account for realistic 3D scene representation. In this study, we propose CoCoGaussian, a \textbf{C}ircle \textbf{o}f \textbf{Co}nfusion-aware \textbf{Gaussian} Splatting that enables precise 3D scene representation using only defocused images. CoCoGaussian addresses the challenge of defocus blur by modeling the Circle of Confusion (CoC) through a physically grounded approach based on the principles of photographic defocus. Exploiting 3D Gaussians, we compute the CoC diameter from depth and learnable aperture information, generating multiple Gaussians to precisely capture the CoC shape. Furthermore, we introduce a learnable scaling factor to enhance robustness and provide more flexibility in handling unreliable depth in scenes with reflective or refractive surfaces. Experiments on both synthetic and real-world datasets demonstrate that CoCoGaussian achieves state-of-the-art performance across multiple benchmarks.
\end{abstract}					
\section{Introduction}
\label{sec:intro}

The emergence of the Neural Radiance Field (NeRF)~\cite{mildenhall2020nerf} has brought significant attention to 3D scene representation for photo-realistic novel view synthesis. They reconstruct 3D scenes from images captured from multiple views and render images of unseen views. However, since NeRF models 3D scenes as implicit neural representations through a ray tracing-based approach, they suffer from inefficient memory usage. In contrast, a rasterization-based method, 3D Gaussian Splatting (3DGS)~\cite{kerbl20233d} explicitly models 3D scenes using independent 3D Gaussians with different 3D means, covariances, opacities, and spherical harmonic coefficients. The tile-based rasterizer of 3DGS applies alpha blending to Gaussian splats sorted by visibility order, enabling fast training, rendering, and efficient memory usage.


Both NeRF~\cite{mildenhall2020nerf} and 3DGS~\cite{kerbl20233d} rely on sharp images to represent 3D scenes accurately, which is a highly ideal assumption. In real-world settings, various factors (\textit{e.g.}, finite depth of field and camera motion blur) can hinder the capture of sharp images, leading to image degradation. Achieving a perfectly sharp image requires every part of the scene to be in focus, namely all-in-focus, which calls for a large depth of field. However, obtaining a large depth of field necessitates using a small aperture, which in turn requires a longer exposure time to allow enough light to enter the camera. During this extended exposure, even slight movements of a handheld camera can introduce motion blur. To prevent motion blur, the aperture needs to be widened, but this comes at the cost of a shallower depth of field. As a result, areas outside the focus plane appear blurred due to defocus. Defocus blur is closely related to the concept of the Circle of Confusion (CoC); as shown in \cref{fig:teaser}, when a subject is positioned away from the focus plane, the radiances from the subject pass through the aperture and are mapped onto the image sensor in a circular pattern, resulting in shallow depth of field and defocus blur~\cite{hecht2012optics}. However, despite these challenges, recent research in 3D scene representation has rarely focused on addressing defocus blur directly. In this paper, we propose a method to represent 3D scenes using only images with defocus blur, tackling the CoC-related challenges posed by real-world image acquisition scenarios.

Recently, there has been increasing interest in reconstructing 3D scenes using only degraded images, with a focus on addressing challenges caused by camera motion blur and defocus blur. Deblur-NeRF~\cite{ma2022deblurnerf} is the first study to take on this task, adopting a blind image deblurring approaches~\cite{whyte2012non,chakrabarti2016neural,srinivasan2017light}. Follow-up studies~\cite{lee2023dp,peng2023pdrf,peng2024bags,lee2024smurf,lee2024crim,wang2023bad,lee2024deblurring} have expanded this approach beyond ray-tracing to rasterization-based methods. However, many of these methods depend heavily on learning-based strategies and fail to incorporate photographic principles that accurately capture how defocus occurs in real-world scenarios. While DoF-NeRF~\cite{wu2022dof} adopts the concept of CoC to model defocus blur in 3D scene representations, it relies on an implicit representation, which leads to slow rendering speeds, making it less suitable for real-time applications.

In this paper, we propose CoCoGaussian, which leverages a physically grounded photographic defocus principles to represent 3D scenes. Our method incorporates an aperture to calculate the CoC diameter based on depth and learnable aperture information using 3D Gaussians, accurately modeling shallow depth of field images. Using the CoC diameter and the 3D Gaussian, we generate multiple Gaussians to form the CoC shape, as illustrated in~\cref{fig:teaser}. However, when objects with refraction or reflection exist in the scene, the depth obtained through the Gaussians may become unreliable. To address this, we propose a method to keep the set of generated Gaussians within the CoC radius, introducing a learnable scaling factor to increase flexibility and reduce dependence on estmiated depth in the CoC modeling. As a result, our model combines physically grounded defocus principles with an adaptive CoC modeling that effectively handles uncertain depths, allowing it to reconstruct sharp 3D scenes using only defocused images. Additionally, our approach has an advantage of learning aperture and focus plane information, enabling dynamic control over depth of field and flexible adjustment of the focus plane. This feature allows for highly customizable scene visualization, adapting to various focus and depth requirements.

To demonstrate the effectiveness of our model, we conduct extensive experiments on the benchmarks: the Deblur-NeRF~\cite{ma2022deblurnerf} dataset and the DoF-NeRF~\cite{wu2022dof} real-world dataset. Our contributions can be summarized as follows:

\begin{itemize}
	\item [$\bullet$] CoCoGaussian models the CoC at the 3D Gaussian level, reconstructing the precise 3D scene and enabling sharp novel view synthesis from defocused images.
	\vspace{1mm}
	\item [$\bullet$] We propose an adaptive learning approach that robustly models the CoC even with unreliable depth information.
	\vspace{1mm}
	\item [$\bullet$] CoCoGaussian enables the customizable depth of field and flexible focus plane adjustment during rendering by learning aperture and focus plane information.
	\vspace{1mm}
	\item [$\bullet$] CoCoGaussian achieves state-of-the-art performance, both quantitatively and qualitatively.
\end{itemize}							
\section{Related Work}
\label{sec:relatedwork}

\subsection{Scene Representations for Novel View Synthesis}

3D scene representation for novel view synthesis has advanced significantly with the advent of neural radiance fields (NeRF)~\cite{mildenhall2020nerf}, a ray tracing-based volumetric rendering method that generates photo-realistic novel view images from multi-view images. However, as an implicit representation method using deep MLPs, NeRF suffers from slow training and rendering speeds. To address this limitation, explicit representation methods such as Plenoxel~\cite{fridovich2022plenoxels}, TensoRF~\cite{chen2022tensorf}, and Instant-NGP~\cite{muller2022instant} have been introduced. More recently, 3D Gaussian Splatting (3DGS)~\cite{kerbl20233d}, a rasterization-based method, has emerged as an alternative to ray tracing, significantly mitigating the speed and memory efficiency constraints of ray tracing-based approaches. NeRF and 3DGS has enabled a wide range of related research, including dynamic scene representation~\cite{pumarola2021dnerf,park2021nerfies,park2021hypernerf,li2022neural3dvideo,tretschk2021nonrigid,li2021nsff}, human avatars~\cite{weng2022humannerf,peng2021animatable,jiang2022neuman}, 3D mesh reconstruction~\cite{wang2021neus,wang2022neuris,li2023neuralangelo,sun2021neuralrecon}, 3D scene representation from sparse-view images~\cite{niemeyer2022regnerf,yang2023freenerf,wang2023sparsenerf,wynn2023diffusionerf}, and 3D scene representation from blurry images~\cite{ma2022deblurnerf,lee2023dp,peng2023pdrf,wang2023bad,lee2024smurf,chen2024deblur,zhao2024bad,peng2024bags,lee2024deblurring,lee2024crim}. In this paper, we propose a method for 3D scene representation from defocused images by exploiting 3DGS.

\subsection{Novel View Synthesis from Blurry Images}

Scene representation methods like NeRF~\cite{mildenhall2020nerf} and 3DGS~\cite{kerbl20233d} require sharp images as input to render photo-realistic images. However, in real-world scenarios, capturing sharp images is challenging, often suffering from image degradation such as camera motion blur and defocus blur. To address this issue, Deblur-NeRF~\cite{ma2022deblurnerf} introduced a 3D ray-based blurring kernel inspired by image blind deblurring~\cite{whyte2012non,chakrabarti2016neural,srinivasan2017light}, which intentionally generates blurry images during training. After training, it renders sharp novel-view images by excluding the trained kernel. Following Deblur-NeRF, various approaches have been proposed to tackle the degradation issue. DP-NeRF~\cite{lee2023dp} introduces a kernel based on the rigid body transformation~\cite{lynch2017modernrobotics} with prior knowledge that object shape remains consistent in static scenes, and PDRF~\cite{peng2023pdrf} proposes a blur estimation method with 2-stage efficient rendering. Recently, methods utilizing 3DGS have been proposed to enable faster rendering. For instance, Deblurring 3DGS~\cite{lee2024deblurring} adjusts Gaussian parameters to generate blurry images during training, and BAGS~\cite{peng2024bags} proposes a blur-agnostic kernel and a blur mask using convolutional neural networks (CNNs). DoF-NeRF~\cite{wu2022dof} has explored modeling defocus blur by leveraging the circle of confusion (CoC) within an implicit neural representation. In this paper, we propose a method that also leverages CoC to model defocus, which enables real-time rendering, while still maintaining high-quality depth-of-field effects.			
\section{Preliminary}
\label{sec:preliminary}

\subsection{3D Gaussian Splatting}

Unlike ray tracing-based methods~\cite{mildenhall2020nerf,chen2022tensorf,barron2021mipnerf}, 3DGS~\cite{kerbl20233d} is built on a rasterization-based approach with differentiable 3D Gaussians. These 3D Gaussians are initialized from a sparse point cloud obtained via a Structure-from-Motion (SfM)~\cite{schonberger2016pixelwise,shan2008highmotiondeblurring} algorithm and are defined as follows: 
\begin{equation} \label{eq:gaussian_represent}
	G(\mathbf{x}) = e^{-\frac{1}{2}(\mathbf{x} - \mu)^{\top}\mathbf{\Sigma}^{-1}(\mathbf{x} - \mu)},
\end{equation}
where $\mathbf{x}\in\mathbb{R}^{3}$ is a point on the Gaussian $G$ centered at the mean vector $\mu\in\mathbb{R}^{3}$ with an covariance matrix $\mathbf{\Sigma}\in\mathbb{R}^{3\times 3}$. The 3D covariance matrix $\mathbf{\Sigma}$ is derived from a learnable scaling vector $\mathbf{s}\in\mathbb{R}^{3}$ and rotation quaternion $\mathbf{q}\in\mathbb{R}^{4}$, from which the scaling matrix $\mathbf{S}\in\mathbb{R}^{3\times 3}$ and rotation matrix $\mathbf{R}\in\mathbb{R}^{3\times 3}$ are obtained and represented as follows:
\begin{equation} \label{eq:covariance_represent}
	\mathbf{\Sigma} = \mathbf{R}\mathbf{S}\mathbf{S}^{\top}\mathbf{R}^{\top}.
\end{equation}
For differentiable splatting~\cite{yifan2019differentiable}, the Gaussians in the 3D world coordinate system are projected into the 2D camera coordinate system. This projection uses the viewing transformation $\mathbf{W}\in\mathbb{R}^{3\times 3}$ and the Jacobian $\mathbf{J}\in\mathbb{R}^{2\times 3}$ of the affined approximation of the projective transformation to derive the 2D covariance $\mathbf{\Sigma}^{\textrm{2D}}\in\mathbb{R}^{2\times2}$: 
\begin{equation} \label{eq:covariance_2d}
	\mathbf{\Sigma}^{\textrm{2D}} = \mathbf{J}\mathbf{W}\mathbf{\Sigma}\mathbf{W}^{\top}\mathbf{J}^{\top}.
\end{equation}
Each Gaussian includes a set of spherical harmonics (SH) coefficients and an opacity value $\alpha$ to represent view-dependent color $\mathbf{c}$. The pixel color $\mathbf{c}_{p}$ is then obtained by applying alpha blending to $\mathcal{N}$ ordered Gaussians:
\begin{equation} \label{eq:alpha_blending}
	\mathbf{c}_{p}=\sum_{i \in \mathcal{N}} \mathbf{c}_i \alpha_i \prod_{j=1}^{i-1}\left(1-\alpha_j\right).
\end{equation}

Our approach, as illustrated in~\cref{fig:teaser}, exploits 3DGS and aims to render defocused images by generating multiple Gaussians to form the CoC for each Gaussian.

\begin{figure*}[t]
	\centering
	\includegraphics[width=\textwidth]{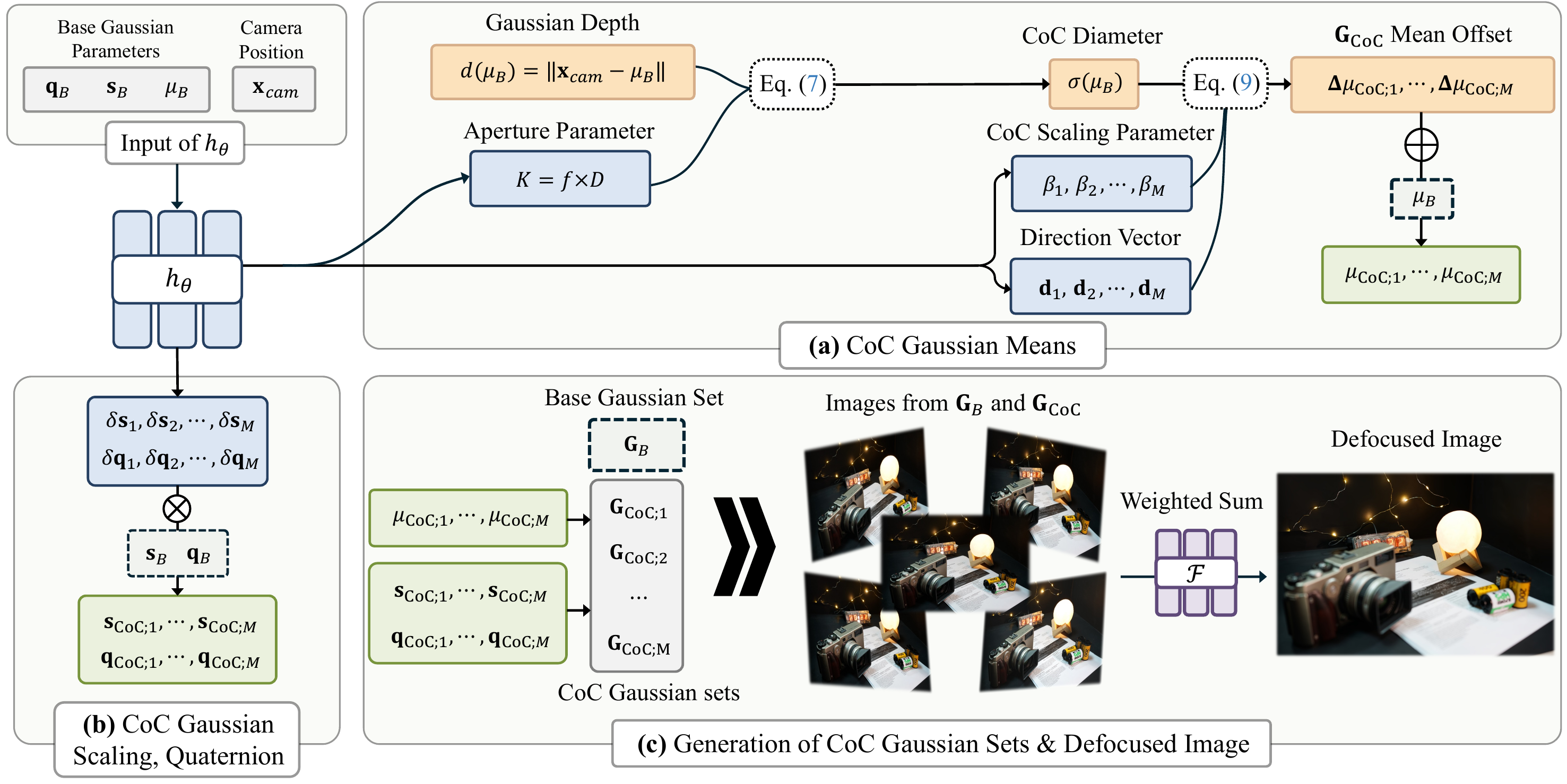}
	\caption{We take the camera position $\mathbf{x}_{cam}$ and the base Gaussian parameters $\mu_{B}$, $\mathbf{s}_{B}$, and $\mathbf{q}_{B}$ as inputs to the MLP $h_{\theta}$, which produces five outputs in total. \textbf{(a)} We set the depth $d({\mu_{B}})$ as the Euclidean distance between $\mathbf{x}_{cam}$ and $\mu_{B}$. Using the output $K$ from $h_{\theta}$, the $d({\mu_{B}})$, and the learnable focus plane $d_{F}$, we apply \cref{eq:approx_coc} to determine the CoC diameter. The diameter, combined with the outputs $\beta$ and $\mathbf{d}$ from $h_{\theta}$, is then used in \cref{eq:delta_coc} to compute the offset values for $\mu_{\textrm{CoC}}$. By adding these offsets to $\mu_{B}$, we obtain the $\mathbf{G}_{\textrm{CoC}}$ means. \textbf{(b)} We obtain $\mathbf{s}_{\textrm{CoC}}$ and $\mathbf{q}_{\textrm{CoC}}$ by applying \cref{eq:scaling,eq:quaternion} to $\delta\mathbf{s}_{\textrm{CoC}}$ and $\delta\mathbf{q}_{\textrm{CoC}}$, the outputs of $h_{\theta}$. \textbf{(c)} Using $\mathbf{\mu}_{\textrm{CoC}}$, $\mathbf{s}_{\textrm{CoC}}$, and $\mathbf{q}_{\textrm{CoC}}$, we get the $\mathbf{G}_{\textrm{CoC}}$. Finally, we rasterize $\mathbf{G}_{\textrm{CoC}}$ along with $\mathbf{G}_{B}$ to produce $(M+1)$ images, and apply a weighted sum to generate the final defocused image.}
	\label{fig:framework}
\end{figure*}

\subsection{3D Scene Blind Deblurring}

Image blind deblurring is a technique for learning an image blurring kernel without the supervision of sharp images. Blurry images are obtained by applying convolution with a learned blurring kernel to sharp images, where the kernel size is fixed as a grid structure at each pixel position. Inspired by this approach, Deblur-NeRF~\cite{ma2022deblurnerf} applies this technique to NeRF, presenting a method to represent a clean 3D scene using only blurry images. Specifically, Deblur-NeRF warps the input ray into multiple rays that constitute the blur, modeling a ray-based sparse kernel. A blurry pixel is obtained by combining the pixel colors of each ray:
\begin{equation} \label{eq:blur_eq}
	\mathbf{c}_{blur} = \sum w_{p}^{i}\mathbf{c}_{p}^{i},~w.r.t.,~\sum w_{p}^{i}=1,
\end{equation}
where $w_{p}$ and $\mathbf{c}_{p}$ denote the weight and pixel color for each corresponding ray. Deblur-NeRF enhances learning efficiency by setting the number of warped rays smaller than the kernel size of 2D convolution, and it designs a deformable kernel by adaptive origins and directions of rays. 

As we adopt a rasterization method based on 3DGS~\cite{kerbl20233d}, we propose an alternative approach: instead of warping rays, we generate 3D Gaussians in the shape of the CoC and render them. In other words, the defocused image is estimated by setting the generated 3D Gaussians, namely CoC Gaussians, as the blurring kernel.			  
\section{Method}
\label{sec:method}

\subsection{CoCoGaussian Framework} \label{sec:cocogaussian_framework}

Our goal is to represent a 3D scene using only defocused multi-view images, leveraging fundamental photographic principles to guide our method. We generate 3D Gaussians to capture a blurring kernel shaped as the CoC. The whole framework of CoCoGaussian is on~\cref{fig:framework}. First, we compute the CoC diameter using the Gaussian depth, derived from the means of the given base 3D Gaussian set $\mathbf{G}_{B}$, which consists of $N$ Gaussians, along with the camera position and aperture information (\cref{sec:coc_from_gaussians}). Next, for the base Gaussian set $\mathbf{G}_{B}$, we generate $M$ CoC Gaussian sets $\mathbf{G}_{\textrm{CoC}}$, resulting in a total of $(M \times N)$ Gaussians to capture the CoC shape. However, when the scene contains objects with refraction or reflection, the Gaussian depth may be unreliable, which can lead to suboptimal CoC diameter. To address this issue, we propose an adaptive CoC Gaussian generation method that minimizes the dependence on depth for $\mathbf{G}_{\textrm{CoC}}$ (\cref{sec:adaptive_gaussian_generation}). Finally, we incorporate adaptive rotation quaternion and scaling vector adjustments, inspired by the Deblurring 3DGS~\cite{lee2024deblurring} approach, to allow flexible learning of $\mathbf{G}_{\textrm{CoC}}$. Furthermore, we apply \cref{eq:blur_eq} to images rendered from $\mathbf{G}_{B}$ and $\mathbf{G}_{\textrm{CoC}}$, using a weighted sum approach~\cite{peng2024bags,lee2024crim} to create the final blurry image. This photography-prior-based approach allows us to accurately represent 3D scenes using only defocused images while adhering to realistic defocus principles.

\subsection{Circle of Confusion from 3D Gaussians} \label{sec:coc_from_gaussians}

To model the CoC, we assume an ideal system that ignores distortions such as lens aberrations. Before generating the CoC Gaussian sets $\mathbf{G}_{\textrm{CoC}}$, we first need to obtain the CoC diameter, which depends on several factors: the aperture diameter $D$, focal length $f$, focus plane $d_{F}$, and depth $p$. The focus plane represents the distance from the camera position to the plane in focus. We define the depth $d(\mathbf{\mu}_{B})$ as the Euclidean distance between the camera position $\mathbf{x}_{cam}\in\mathbb{R}^{3}$ and the means of the base Gaussian set $\mathbf{\mu}_{B}\in\mathbb{R}^{N\times3}$. Exploiting these values, we express the CoC diameter $\sigma(\mu_{B})$ as follows~\cite{hecht2012optics}:
\begin{equation} \label{eq:circle_of_confusion}
	\sigma(\mathbf{\mu}_{B}) = fD\times \frac{|d(\mathbf{\mu}_{B}) - d_{F}|}{d(\mathbf{\mu}_{B})(d_{F} - f)},
\end{equation}
where $d_{F}$ is set to a learnable parameter and initialized as the average distance between $\mathbf{x}_{cam}$ and the points in the SfM point cloud to ensure stable training. Note that the focus plane $d_{F}$ is specific to each input image. Additionally, because the focal length $f$ is usually much smaller than the focus plane $d_{F}$, we modify \cref{eq:circle_of_confusion} as follows:
\begin{equation} \label{eq:approx_coc}
	\sigma(\mathbf{\mu}_{B}) \approx K \times \left| \frac{1}{d(\mathbf{\mu}_{B})} - \frac{1}{d_{F}} \right|,
\end{equation}
where the product of the focal length and aperture diameter is represented as a single learnable scalar $K = f \times D$~\cite{wu2022dof}.

We design a simple MLP $h_{\theta}$ parameterized by $\theta$ to estimate the scalar $K$. Then, the aperture parameter $K$ is obtained by passing these features as inputs to $h_{\theta}$, along with the scaling vectors $\mathbf{s}_{B}\in\mathbb{R}^{N\times 3}$ and rotation quaternions $\mathbf{q}_{B}\in\mathbb{R}^{N\times 4}$ of $\mathbf{G}_{B}$. This approach allows us to obtain the CoC diameter $\sigma(\mathbf{\mu}_{B})$ for the $\mathbf{G}_{B}$, marking the first step in modeling $\mathbf{G}_{\textrm{CoC}}$.

\subsection{Adaptive CoC Gaussian Generation} \label{sec:adaptive_gaussian_generation}

After obtaining the CoC diameters, we use these diameters to generate $M$ CoC Gaussian sets $\mathbf{G}_{\textrm{CoC}}$. For the Gaussian parameters of $\mathbf{G}_{\textrm{CoC}}$, we set the spherical harmonics (SH) coefficients and opacity values identical to those of $\mathbf{G}_{B}$, while varying only the means $\mathbf{\mu}$, scaling vectors $\mathbf{s}$, and rotation quaternions $\mathbf{q}$. The means of each CoC Gaussian set are created by adding offsets $\Delta\mathbf{\mu}_{\textrm{CoC}}\in\mathbb{R}^{M\times N\times 3}$ to $\mathbf{\mu}_{B}$, defined as $\mathbf{\mu}_{\textrm{CoC}} = \mathbf{\mu}_{B} + \Delta\mathbf{\mu}_{\textrm{CoC}}$. Since $\Delta\mathbf{\mu}_{\textrm{CoC}}$ are 3D vector sets, we obtain unit vector sets $\mathbf{d}\in\mathbb{R}^{M\times N\times 3}$ representing the directions from $\mathbf{G}_{B}$ to $\mathbf{G}_{\textrm{CoC}}$, which are additional outputs from the MLP $h_{\theta}$ discussed in the previous section. Moreover, since $\mathbf{G}_{\textrm{CoC}}$ consist of a total of $M$ Gaussian sets, The direction vectors $\mathbf{d}$ consist of $M$ instances:
\begin{equation}
	\Delta\mathbf{\mu}_{\textrm{CoC};m} = \frac{\sigma(\mathbf{\mu}_{B})}{2}\mathbf{d}_{m},~~where~~ m \in M.
\end{equation}

However, theses offsets $\Delta\mathbf{\mu}_{\textrm{CoC}}$ presents two potential issues. First, this approach places all $\mathbf{G}_{\textrm{CoC}}$ only on the boundary of the CoC, which limits its ability to fully capture the defocus effect. Modeling the entire CoC, including its interior, would allow for a more accurate and flexible representation of defocus blur. The second issue arises when there are refractive or reflective surfaces in the scene, as these subjects can lead to incorrect optimization of the Gaussian means, making the CoC diameter obtained through \cref{eq:approx_coc} unreliable. To address these problems, we introduce simple learnable CoC scaling parameters $\beta \in (0,1]$, which are additional outputs of $h_{\theta}$. The parameter $\beta$ is learned within the range of $0$ to $1$, ensuring that $\mathbf{\mu}_{\textrm{CoC}}$ resides within the CoC boundary. Additionally, even when depth is inaccurately measured and the predicted CoC diameter is larger than the actual size, $\beta$ adaptively scales it down. Thus, the offset $\Delta\mathbf{\mu}_{\textrm{CoC}}$ is modified by applying $\beta$ as follows:
\begin{equation}\label{eq:delta_coc}
	\Delta \mathbf{\mu}_{\textrm{CoC};m} = \frac{\sigma(\mathbf{\mu}_{B})}{2}\beta_{m}\mathbf{d}_{m}.
\end{equation}

Additionally, similar to Deblurring 3DGS~\cite{lee2024deblurring}, we introduce scaling factors $\delta\mathbf{s}_{\textrm{CoC}}\in\mathbb{R}^{M\times N\times 3}$ and $\delta\mathbf{q}_{\textrm{CoC}}\in\mathbb{R}^{M\times N\times 4}$, which allow flexible learning of the scaling vectors $\mathbf{s}_{\textrm{CoC}}$ and rotation quaternions $\mathbf{q}_{\textrm{CoC}}$ of $\mathbf{G}_{\textrm{CoC}}$. These factors are also the outputs of $h_{\theta}$. The parameters $\delta\mathbf{s}_{\textrm{CoC}}$ and $\delta\mathbf{q}_{\textrm{CoC}}$ are multiplied by the corresponding parameters of the base Gaussian set $\mathbf{G}_{B}$, specifically $\mathbf{s}_{B}$ and $\mathbf{q}_{B}$, and are expressed as follows: 
\begin{equation} \label{eq:scaling}
	\mathbf{s}_{\textrm{CoC};m} = \mathbf{s}_{B} \times\delta\mathbf{s}_{\textrm{CoC};m},
\end{equation}
\begin{equation}\label{eq:quaternion}
	\mathbf{q}_{\textrm{CoC};m} = \mathbf{q}_{B} \times\delta\mathbf{q}_{\textrm{CoC};m},
\end{equation}
where the scaling parameters are constrained by the fixed values $\delta\mathbf{s}_{\max}$ and $\delta\mathbf{q}_{\max}$: $\delta\mathbf{s}_{\textrm{CoC}} \in [1, \delta\mathbf{s}_{\max}]$ and $\delta\mathbf{q}_{\textrm{CoC}} \in [1, \delta\mathbf{q}_{\max}]$.

We obtain a total of $M$ means of CoC Gaussian sets $\mathbf{\mu}_{\textrm{CoC}}$, scaling vectors $\mathbf{s}_{\textrm{CoC}}$, and rotation quaternion $\mathbf{q}_{\textrm{CoC}}$. They are combined with the opacity and SH coefficients of $\mathbf{G}_{B}$, generate the $M$ Gaussian sets $\mathbf{G}_{\textrm{CoC}}$:
\begin{equation}
	\mathbf{x}_{\textrm{CoC};m},~\mathbf{s}_{\textrm{CoC};m},~\mathbf{q}_{\textrm{CoC};m},~\alpha_{B},~\textrm{SH}_{B} \rightarrow \mathbf{G}_{\textrm{CoC};m}.
\end{equation}
We rasterize the resulting $M$ Gaussian sets along with the base Gaussian set, producing $(M+1)$ images. Then, using the weighted sum approach from \cref{eq:blur_eq}, as detailed in the following section, we obtain the final blurry image.

\subsection{Optimization} \label{sec:optimization}

\paragraph{Weighted Sum.}To apply \cref{eq:blur_eq} to the $(M+1)$ images $\mathbf{I}$ rasterized from $\mathbf{G}_{B}$ and $\mathbf{G}_{\textrm{CoC}}$, we adopt the methods of recent studies~\cite{peng2024bags,lee2024crim}. We use a shallow CNN $\mathcal{F}$ to compute pixel-wise weights $\mathcal{W}$ for these images, applying a softmax function to ensure that the weights for each pixel sum to 1. Then, we multiply each image by its corresponding weight and sum the results to obtain the final defocused image $\mathcal{I}_{blur}$: 
\begin{equation}
	\mathcal{I}_{blur} = \sum_{m=1}^{M+1} \mathcal{I}_{m}\mathcal{W}_{m},~where~\mathcal{W}=\textrm{softmax}(\mathcal{F}(\mathbf{I})),
\end{equation}
where $\mathcal{I}_{m}$ represents the image rasterized from $\mathbf{G}_{\textrm{CoC};m}$, and $\mathcal{I}_{M+1}$ represents the image rasterized from $\mathbf{G}_{B}$.

\paragraph{Objective.}We minimize the $\mathcal{L}_{1}$ loss and D-SSIM loss $\mathcal{L}_{\textrm{D-SSIM}}$ between the ground truth defocused image and the output defocused image. The $\mathcal{L}_{1}$ loss reduces the pixel-wise differences between images, while the D-SSIM loss minimizes structural differences between them. The final objective $\mathcal{L}_{rgb}$ is defined as follows: 
\begin{equation}
	\mathcal{L}_{rgb} = (1 - \lambda)\mathcal{L}_{1} + \lambda\mathcal{L}_{\textrm{D-SSIM}},
\end{equation}
where we use $\lambda=0.3$ for all experiments.

\section{Experiments}
\label{sec:experiments}

\begin{table}[!t] 
	\begin{center}
		\caption{Comparisons on Deblur-NeRF synthetic and real-world scene dataset. We evaluate the performance on three metrics (PSNR, SSIM, LPIPS). ``*'' denotes the results obtained by reproducing the released code. The \colorbox{best!25}{orange} and \colorbox{second!35}{yellow} cells respectively indicate the highest and second-highest value.}
		\resizebox{\columnwidth}{!}{
			\centering
			\setlength{\tabcolsep}{1pt}
			\renewcommand{\arraystretch}{1.1}
			\scriptsize
			\begin{tabular}{l||c|c|c|c|c|c}
				\toprule 
				
				\multirow{2}{*}{Methods} 			   	& \multicolumn{3}{c|}{~Synthetic Scene~\cite{ma2022deblurnerf}~ }  	   & \multicolumn{3}{c}{~Real-World Scene~\cite{ma2022deblurnerf}~}  	\\ \cmidrule{2-7}
				&~PSNR$\uparrow$~    &~SSIM$\uparrow$~    &~LPIPS$\downarrow$~ &~PSNR$\uparrow$~    &~SSIM$\uparrow$~    &~LPIPS$\downarrow$~     	\\ \midrule \midrule
				Naive NeRF~\cite{mildenhall2020nerf}                            				 & 25.95    & 0.7791   & 0.2303 		& 22.40       & 0.6661      & 0.2310 	 	\\
				3DGS~\cite{kerbl20233d}                           						& 25.11    & 0.7476   & 0.2148 			& 23.38       & 0.6655      & 0.3140      	\\ \midrule
				Deblur-NeRF~\cite{ma2022deblurnerf}                          			& 28.37    & 0.8527   & 0.1188 		 & 23.47       & 0.7199      & 0.1207 	\\
				PDRF-10~\cite{peng2023pdrf}                         					   & 30.08   & 0.8931   & 0.1101 		 		  & \cellcolor{best!25}23.85       & 0.7382      & 0.1746        				\\
				DP-NeRF~\cite{lee2023dp}                        						  & 29.33    & 0.8713  & 0.0987 		& 23.67   	 & 0.7299      & 0.1082   \\ \midrule
				Deblurring 3DGS*~\cite{lee2024deblurring}~                        				& 28.90    & 0.8912   & 0.1052 		 & 23.54       & 0.7383      & 0.1232 	\\ 
				BAGS*~\cite{peng2024bags}                          				& \cellcolor{second!35}30.65    & \cellcolor{second!35}0.9128  & \cellcolor{second!35}0.0631 		 & 23.48       & \cellcolor{second!35}0.7408      & \cellcolor{second!35}0.0962 	\\ \midrule \midrule
				\textbf{Ours}											                        & \cellcolor{best!25}30.84    & \cellcolor{best!25}0.9212   & \cellcolor{best!25}0.0478 		 & \cellcolor{second!35}23.70       & \cellcolor{best!25}0.7531      & \cellcolor{best!25}0.0825	\\ \bottomrule
			\end{tabular}
		}
		\label{tab:comparison_deblur}
	\end{center}
	\vspace{-3mm}
\end{table}

\begin{table}[!t]
	\centering
	\caption{Comparisons on DoF-NeRF real-world scene dataset.}
	\resizebox{0.9\columnwidth}{!}{
		\setlength{\tabcolsep}{7pt} 
		\renewcommand{\arraystretch}{1.0}
		\scriptsize 
		\begin{tabular}{l||c|c|c}
			\toprule
			\multirow{2}{*}{Methods} & \multicolumn{3}{c}{~DoF-NeRF Real-World Scene~\cite{wu2022dof}~} \\ \cmidrule{2-4}
			& PSNR$\uparrow$ & SSIM$\uparrow$ & LPIPS$\downarrow$ \\ \midrule \midrule
			3DGS~\cite{kerbl20233d} & 25.72 & 0.8291 & 0.1817 \\ \midrule
			DP-NeRF~\cite{lee2023dp}~ &  26.80 & 0.8145  & 0.1790  \\ \midrule
			Deblurring 3DGS~\cite{lee2024deblurring}~ & 26.58 & 0.8386 & 0.1708 \\ 
			BAGS~\cite{peng2024bags} & \cellcolor{second!35}29.87 & \cellcolor{second!35}0.8816 & \cellcolor{second!35}0.1100 \\ \midrule \midrule
			\textbf{Ours} & \cellcolor{best!25}30.14 & \cellcolor{best!25}0.9127 & \cellcolor{best!25}0.0701 \\ \bottomrule
		\end{tabular}
	}
	\label{tab:comparison_dof}
\end{table}

\paragraph{Datasets.}We evaluate our method on three different datasets: the Deblur-NeRF~\cite{ma2022deblurnerf} synthetic dataset, the Deblur-NeRF real-world dataset, and the DoF-NeRF~\cite{wu2022dof} real-world dataset. The Deblur-NeRF synthetic dataset consists of 5 scenes, each containing defocused images generated using the built-in function of Blender~\cite{blender}. The Deblur-NeRF real-world dataset includes 10 scenes, with images captured by enlarging the aperture on a Canon EOS RP. The DoF-NeRF real-world dataset comprises 7 scenes, each containing 20 to 30 image triplets. Each triplet includes an all-in-focus image, along with two defocused images focused on the background and foreground. Since DoF-NeRF does not provide training code, we use half of the training defocused images with a foreground focus and the other half with a background focus. All camera poses and initial point clouds are obtained through COLMAP~\cite{schonberger2016pixelwise,shan2008highmotiondeblurring}.


\subsection{Rendering Results}

\paragraph{Quantitative Results.} We quantitatively evaluate our CoCoGaussian using the following three metrics: peak signal-to-noise ratio (PSNR), structural similarity index measure (SSIM)~\cite{wang2004image}, and learned perceptual image patch similarity (LPIPS)~\cite{zhang2018lpips}. We compare our approach to several state-of-the-art methods, including both ray-tracing~\cite{ma2022deblurnerf,lee2023dp,peng2023pdrf} and rasterization-based methods~\cite{peng2024bags,lee2024deblurring}. Quantitative results are shown in \cref{tab:comparison_deblur} and \cref{tab:comparison_dof}, where our approach achieves the best performance across all metrics except for PSNR on the Deblur-NeRF real-world dataset. Note that we report the reproduced performance scores for Deblurring 3DGS~\cite{lee2024deblurring} and BAGS~\cite{peng2024bags} using their official codes. The Deblur-NeRF real-world dataset inherently presents challenges, as there are illumination differences between the sharp and defocused images, leading to relatively lower PSNR score. We discuss this issue in the \textbf{appendix}. For the DoF-NeRF dataset, CoCoGaussian achieves outstanding performance across all metrics, with an approximately 0.04 reduction in LPIPS compared to BAGS. Since the DoF-NeRF dataset has a higher resolution than the Deblur-NeRF dataset, this performance drop in BAGS is likely due to its requirement for manual adjustment of blurring kernel size, highlighting an intrinsic limitation. Conversely, Deblurring 3DGS shows minimal improvement over 3DGS, which can also be attributed to its limited kernel size relative to the higher resolution. The details regarding the time complexity of our model are provided in the \textbf{appendix}.

\begin{figure*}[t]
	\centering
	\includegraphics[width=\textwidth]{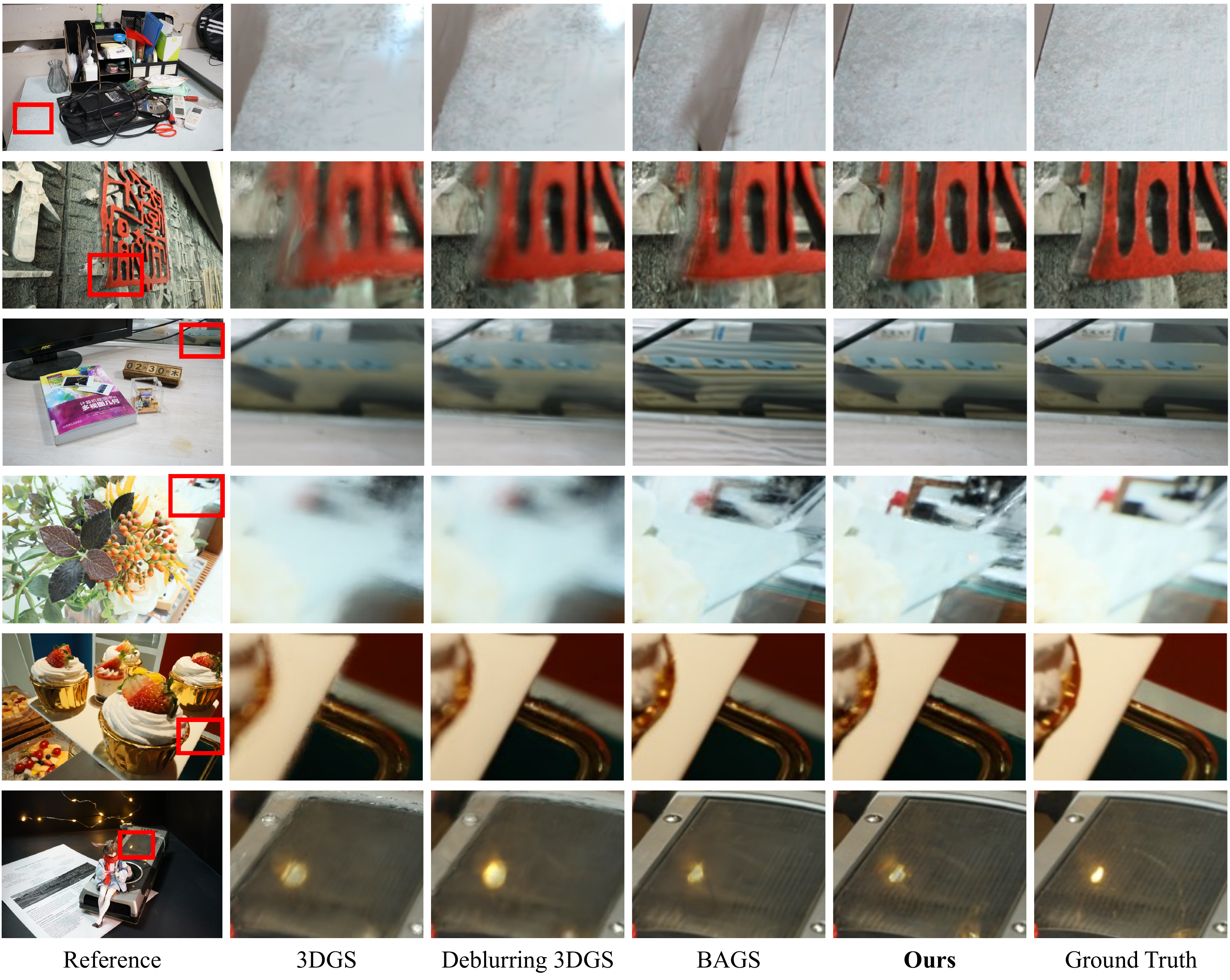}
	\caption{Qualitative comparison on the Deblur-NeRF and DoF-NeRF datasets.}
	\vspace{-5mm}
	\label{fig:comparison}
\end{figure*}

\paragraph{Qualitative Results.} For qualitative evaluation, we present the rendering results of 3DGS~\cite{kerbl20233d}, Deblurring 3DGS~\cite{lee2024deblurring}, and BAGS~\cite{peng2024bags} in \cref{fig:comparison}. Our method demonstrates higher-fidelity results compared to others. Specifically, in the fourth row of our results, the white line on the rightside is rendered more sharply, and in the fifth row, the detailed quality of the graphic card appears superior to that of BAGS. Additionally, more CoC visualizations similar to \cref{fig:teaser} \textbf{(b)}, are in the \textbf{appendix}.

\subsection{Ablation Study}

To validate the effectiveness of our proposed methods, we perform a series of ablation studies, with all experiments conducted on the DoF-NeRF real-world dataset. We chose this dataset as it is based on real captured images and does not present the illumination issues found in the Deblur-NeRF real-world dataset, ensuring more reliable evaluation. The results are summarized in \cref{tab:ablation}, where the baseline denotes naive 3DGS~\cite{kerbl20233d}. Additional ablative experiments with other factors are in the \textbf{appendix}.

\paragraph{Circle of Confusion.}To demonstrate the impact of our core concept, CoC, we conduct an experiment that excludes the CoC from the model, retaining only $\beta_{m}$ and $\mathbf{d}_{m}$. This configuration yields significantly lower performance than the full model, indicating an over-reliance on parametric learning. The result suggests that without the CoC, which leverages depth and aperture information, the means of the generated Gaussian sets are not optimally arranged, leading to overfitting.

\paragraph{CoC Direction Vector.}In this experiment, we retain the CoC concept but exclude the learnable direction vectors $\mathbf{d}_{m}$. Instead of using learnable vectors, we use fixed direction vectors, arranging $M$ CoC Gaussians evenly in a circular pattern on a plane perpendicular to the vector between the camera position and the base Gaussian. This configuration slightly underperforms compared to the full model, as the fixed directions restrict optimal Gaussian position, preventing adaptation to the most reliable configuration.

\paragraph{CoC Scaling Factor.}In this setup, we include the learnable direction vector while omitting the CoC scaling factor $\beta_{m}$. Although the performance is slightly better than when the direction vector is excluded, it remains lower than that of the full model. Without the CoC scaling factor, CoC Gaussian generation becomes overly dependent on depth, leading to errors in CoC diameter calculation when depth values are inaccurate. Including the CoC scaling factor enables the full model to handle uncertain depth more robustly, improving performance. Related visualization results are in the \textbf{appendix}.

\begin{figure*}[t]
	\centering
	\includegraphics[width=\textwidth]{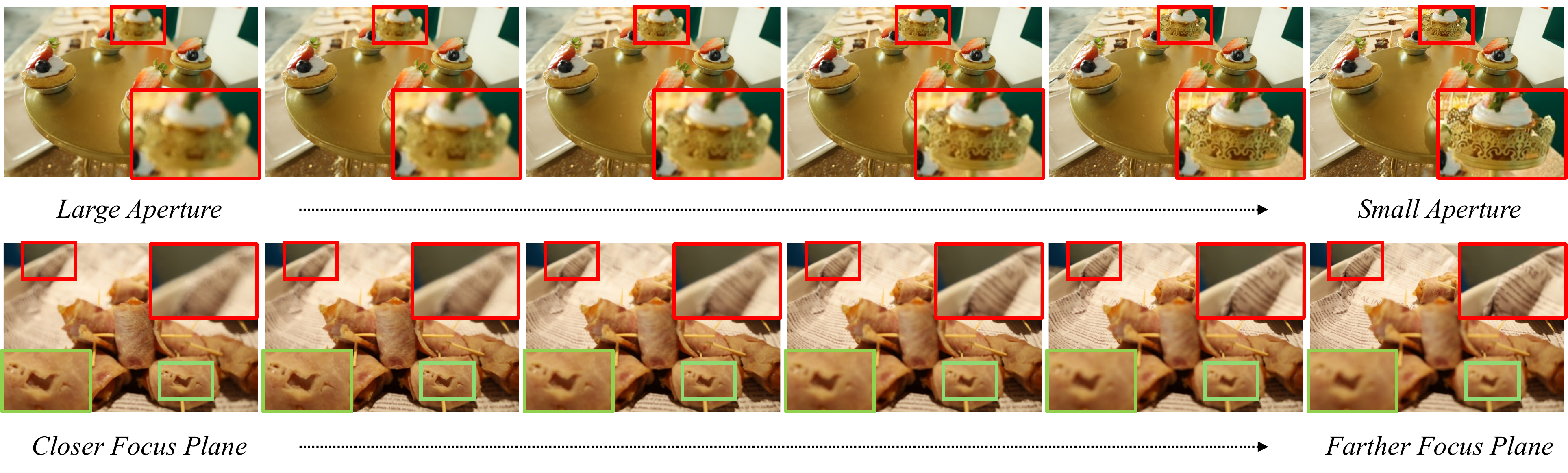}
	\vspace{-7mm}
	\caption{Visualization of Aperture parameter and Focus Plane Customization. The top row of images decreases the aperture parameter $K$ from left to right, while the bottom row moves the focus plane $d_{F}$ further from the camera from left to right.}
	\vspace{-3mm}
	\label{fig:customize}
\end{figure*}

\paragraph{Aperture Parameter.}The aperture parameter $K$ includes the aperture diameter and focal length, and it plays a critical role in determining the size of the CoC. Without this parameter, there is an upper limit to the CoC diameter calculation, making accurate computation challenging. Experimental results indicate that although performance is somewhat lower than the full model, it is improved over the baseline. This suggests that while the model is unable to determine precise CoC size without $K$, it still learns the focus plane and captures the in-focus region accurately.

\subsection{Experiments on All-in-Focus Images.}

We conduct novel view synthesis experiments using all-in-focus images to verify the generalization performance of CoCoGaussian. We utilize the NeRF-LLFF~\cite{mildenhall2019local,mildenhall2020nerf} dataset, which contains real-world images. As shown in~\cref{tab:nerf_llff}, our approach outperforms the baseline 3DGS~\cite{kerbl20233d} across all metrics. In practice, when the aperture is very small, radiance from a point on the subject forms a tiny CoC on the image sensor after passing through the aperture, even if the subject is slightly away from the focus plane. Because the CoC is so small, the resulting image is not perceived as defocused. By modeling this small circle, which is difficult to detect with the human eye, our approach achieves higher performance than the baseline 3DGS. Related CoC visualizations are in the \textbf{appendix}.

\begin{table}[t] 
	\begin{center}
		\caption{Ablation Results of CoCoGaussian.}
		\vspace{-2mm}
		\resizebox{1\columnwidth}{!}{
			\centering
			\setlength{\tabcolsep}{7pt} 
			\begin{tabular}{l||c|c|c}
				\toprule 
				Methods					 &~~PSNR$\uparrow$~~   &~~SSIM$\uparrow$~~    &~~LPIPS$\downarrow$~~     	\\ \midrule \midrule
				Baseline~\cite{kerbl20233d}                           						 & 25.72    & 0.8291   & 0.1817	\\ \midrule
				w/o CoC																							&28.29	& 0.8778	& 0.0927	\\
				w/o CoC direction vector~																	& 28.91	& 0.8893	& 0.0896	\\
				w/o CoC scaling											                         		& 29.46 & 0.9058  & 0.0793 \\ 
				w/o aperture parameter~											                         		& 27.37 & 0.8510  & 0.1113  \\ \midrule
				Ours Full											                         		& \cellcolor{best!25}30.14    & \cellcolor{best!25}0.9127   & \cellcolor{best!25}0.0701	\\ \bottomrule
			\end{tabular}
		}
		\label{tab:ablation}
		\vspace{-9mm}
	\end{center}
\end{table}

\subsection{Customizable Depth of Field and Focus Plane}

Customizing the depth of field and focus settings is essential in various applications, such as augmented reality (AR) and virtual reality (VR), where control over depth cues and focus effects can greatly enhance visual realism and viewer engagement. Since our approach incorporates the aperture parameter $K$ in CoC modeling, it enables customization of both CoC size and depth of field by adjusting this parameter. The top row in \cref{fig:customize} visualizes increasing depth of field by gradually reducing $K$ from left to right. As the focus plane is positioned at a shallow depth, the model consistently captures focus in this region. However, in the leftmost image, the large aperture causes the CoC size to increase at deeper depths, resulting in defocus blur.

Additionally, our model can customize the focus plane of defocused images, as shown in the bottom row of \cref{fig:customize}, where the focus plane shifts from shallow depth on the left to deeper depth on the right. The shallow regions are in focus on the leftmost image, while deeper areas remain out of focus, and the opposite holds for the rightmost image. This level of control over focus and depth effects is essential for achieving realistic depth cues. Therefore, CoCoGaussian not only demonstrates superior capability in synthesizing sharp novel view images but also provides enhanced flexibility in adjusting aperture and focus plane settings.

\begin{table}[t] 
	\begin{center}
		\caption{Results on NeRF-LLFF~\cite{mildenhall2020nerf,mildenhall2019local} Dataset.}
		\vspace{-2mm}
		\resizebox{0.7\columnwidth}{!}{
			\centering
			\setlength{\tabcolsep}{7pt} 
			\begin{tabular}{l||c|c|c}
				\toprule 
				Methods					 &~~PSNR$\uparrow$~~   &~~SSIM$\uparrow$~~    &~~LPIPS$\downarrow$~~     	\\ \midrule \midrule
				3DGS~\cite{kerbl20233d}                           						 & 27.10    & 0.8619   & 0.0569	\\
				Ours											                         		& \cellcolor{best!25}27.76    & \cellcolor{best!25}0.8738   & \cellcolor{best!25}0.0372	\\ \bottomrule
			\end{tabular}
		}
		\label{tab:nerf_llff}
		\vspace{-8mm}
	\end{center}
\end{table}			%
\section{Limitation and Future Work}
\label{sec:limitation}

While our model demonstrates strong qualitative and quantitative performance, there is potential for further optimization in the adaptive CoC scaling factor. To ensure training stability, we currently limit the factor size, enabling adaptive optimization mainly when the CoC diameter is overestimated. However, future improvements could enhance adaptability to handle cases where the CoC diameter is underestimated, further improving depth handling in complex scenes without compromising stability.				%
\section{Conclusion}
\label{sec:conclusion}
In this work, we propose CoCoGaussian, a 3D scene representation method that effectively leverages photographic defocus principles to handle defocused image inputs. CoCoGaussian accurately models defocus blur by constructing the CoC through 3D Gaussians and learnable aperture parameters. We introduce an adaptive CoC Gaussian generation method to address the challenges posed by uncertain depth, making our model robust to reflective or refractive objects. Our approach demonstrates the feasibility and effectiveness of representing 3D scenes from defocused images, advancing possibilities for applications in real-world image synthesis.

\paragraph{Acknowledgements.}
This project was supported by the NAVER Cloud Corporation and the National Research Foundation of Korea (NRF) grant funded by the Korea government (MSIT) (No. RS-2024-00340745). 

{
	\small
	\bibliographystyle{ieeenat_fullname}
	\bibliography{main}

\begin{thebibliography}{47}
\providecommand{\natexlab}[1]{#1}
\providecommand{\url}[1]{\texttt{#1}}
\expandafter\ifx\csname urlstyle\endcsname\relax
  \providecommand{\doi}[1]{doi: #1}\else
  \providecommand{\doi}{doi: \begingroup \urlstyle{rm}\Url}\fi

\bibitem[Barron et~al.(2021)Barron, Mildenhall, Tancik, Hedman, Martin-Brualla,
  and Srinivasan]{barron2021mipnerf}
Jonathan~T Barron, Ben Mildenhall, Matthew Tancik, Peter Hedman, Ricardo
  Martin-Brualla, and Pratul~P Srinivasan.
\newblock Mip-nerf: A multiscale representation for anti-aliasing neural
  radiance fields.
\newblock In \emph{Proceedings of the IEEE/CVF International Conference on
  Computer Vision}, pages 5855--5864, 2021.

\bibitem[Chakrabarti(2016)]{chakrabarti2016neural}
Ayan Chakrabarti.
\newblock A neural approach to blind motion deblurring.
\newblock In \emph{Computer Vision--ECCV 2016: 14th European Conference,
  Amsterdam, The Netherlands, October 11-14, 2016, Proceedings, Part III 14},
  pages 221--235. Springer, 2016.

\bibitem[Chen et~al.(2022)Chen, Xu, Geiger, Yu, and Su]{chen2022tensorf}
Anpei Chen, Zexiang Xu, Andreas Geiger, Jingyi Yu, and Hao Su.
\newblock Tensorf: Tensorial radiance fields.
\newblock In \emph{European Conference on Computer Vision}, pages 333--350.
  Springer, 2022.

\bibitem[Chen and Liu(2024)]{chen2024deblur}
Wenbo Chen and Ligang Liu.
\newblock Deblur-gs: 3d gaussian splatting from camera motion blurred images.
\newblock \emph{Proceedings of the ACM on Computer Graphics and Interactive
  Techniques}, 7\penalty0 (1):\penalty0 1--15, 2024.

\bibitem[Community(2018)]{blender}
Blender~Online Community.
\newblock \emph{Blender - a 3D modelling and rendering package}.
\newblock Blender Foundation, Stichting Blender Foundation, Amsterdam, 2018.

\bibitem[Fridovich-Keil et~al.(2022)Fridovich-Keil, Yu, Tancik, Chen, Recht,
  and Kanazawa]{fridovich2022plenoxels}
Sara Fridovich-Keil, Alex Yu, Matthew Tancik, Qinhong Chen, Benjamin Recht, and
  Angjoo Kanazawa.
\newblock Plenoxels: Radiance fields without neural networks.
\newblock In \emph{Proceedings of the IEEE/CVF Conference on Computer Vision
  and Pattern Recognition}, pages 5501--5510, 2022.

\bibitem[Hecht(2012)]{hecht2012optics}
Eugene Hecht.
\newblock \emph{Optics}.
\newblock Pearson Education India, 2012.

\bibitem[Jiang et~al.(2022)Jiang, Yi, Samei, Tuzel, and
  Ranjan]{jiang2022neuman}
Wei Jiang, Kwang~Moo Yi, Golnoosh Samei, Oncel Tuzel, and Anurag Ranjan.
\newblock Neuman: Neural human radiance field from a single video.
\newblock In \emph{European Conference on Computer Vision}, pages 402--418.
  Springer, 2022.

\bibitem[Kerbl et~al.(2023)Kerbl, Kopanas, Leimk{\"u}hler, and
  Drettakis]{kerbl20233d}
Bernhard Kerbl, Georgios Kopanas, Thomas Leimk{\"u}hler, and George Drettakis.
\newblock 3d gaussian splatting for real-time radiance field rendering.
\newblock \emph{ACM Transactions on Graphics}, 42\penalty0 (4):\penalty0 1--14,
  2023.

\bibitem[Lee et~al.(2024{\natexlab{a}})Lee, Lee, Sun, Ali, and
  Park]{lee2024deblurring}
Byeonghyeon Lee, Howoong Lee, Xiangyu Sun, Usman Ali, and Eunbyung Park.
\newblock Deblurring 3d gaussian splatting.
\newblock \emph{arXiv preprint arXiv:2401.00834}, 2024{\natexlab{a}}.

\bibitem[Lee et~al.(2023)Lee, Lee, Shin, and Lee]{lee2023dp}
Dogyoon Lee, Minhyeok Lee, Chajin Shin, and Sangyoun Lee.
\newblock Dp-nerf: Deblurred neural radiance field with physical scene priors.
\newblock In \emph{Proceedings of the IEEE/CVF Conference on Computer Vision
  and Pattern Recognition}, pages 12386--12396, 2023.

\bibitem[Lee et~al.(2024{\natexlab{b}})Lee, Kim, Lee, Cho, and
  Lee]{lee2024crim}
Junghe Lee, Donghyeong Kim, Dogyoon Lee, Suhwan Cho, and Sangyoun Lee.
\newblock Crim-gs: Continuous rigid motion-aware gaussian splatting from motion
  blur images.
\newblock \emph{arXiv preprint arXiv:2407.03923}, 2024{\natexlab{b}}.

\bibitem[Lee et~al.(2024{\natexlab{c}})Lee, Lee, Lee, Kim, and
  Lee]{lee2024smurf}
Jungho Lee, Dogyoon Lee, Minhyeok Lee, Donghyung Kim, and Sangyoun Lee.
\newblock Smurf: Continuous dynamics for motion-deblurring radiance fields.
\newblock \emph{arXiv preprint arXiv:2403.07547}, 2024{\natexlab{c}}.

\bibitem[Li et~al.(2022)Li, Slavcheva, Zollhoefer, Green, Lassner, Kim,
  Schmidt, Lovegrove, Goesele, Newcombe, et~al.]{li2022neural3dvideo}
Tianye Li, Mira Slavcheva, Michael Zollhoefer, Simon Green, Christoph Lassner,
  Changil Kim, Tanner Schmidt, Steven Lovegrove, Michael Goesele, Richard
  Newcombe, et~al.
\newblock Neural 3d video synthesis from multi-view video.
\newblock In \emph{Proceedings of the IEEE/CVF Conference on Computer Vision
  and Pattern Recognition}, pages 5521--5531, 2022.

\bibitem[Li et~al.(2021)Li, Niklaus, Snavely, and Wang]{li2021nsff}
Zhengqi Li, Simon Niklaus, Noah Snavely, and Oliver Wang.
\newblock Neural scene flow fields for space-time view synthesis of dynamic
  scenes.
\newblock In \emph{Proceedings of the IEEE/CVF Conference on Computer Vision
  and Pattern Recognition}, pages 6498--6508, 2021.

\bibitem[Li et~al.(2023)Li, M{\"u}ller, Evans, Taylor, Unberath, Liu, and
  Lin]{li2023neuralangelo}
Zhaoshuo Li, Thomas M{\"u}ller, Alex Evans, Russell~H Taylor, Mathias Unberath,
  Ming-Yu Liu, and Chen-Hsuan Lin.
\newblock Neuralangelo: High-fidelity neural surface reconstruction.
\newblock In \emph{Proceedings of the IEEE/CVF Conference on Computer Vision
  and Pattern Recognition}, pages 8456--8465, 2023.

\bibitem[Lynch and Park(2017)]{lynch2017modernrobotics}
Kevin~M Lynch and Frank~C Park.
\newblock \emph{Modern robotics}.
\newblock Cambridge University Press, 2017.

\bibitem[Ma et~al.(2022)Ma, Li, Liao, Zhang, Wang, Wang, and
  Sander]{ma2022deblurnerf}
Li Ma, Xiaoyu Li, Jing Liao, Qi Zhang, Xuan Wang, Jue Wang, and Pedro~V Sander.
\newblock Deblur-nerf: Neural radiance fields from blurry images.
\newblock In \emph{Proceedings of the IEEE/CVF Conference on Computer Vision
  and Pattern Recognition}, pages 12861--12870, 2022.

\bibitem[Mildenhall et~al.(2019)Mildenhall, Srinivasan, Ortiz-Cayon, Kalantari,
  Ramamoorthi, Ng, and Kar]{mildenhall2019local}
Ben Mildenhall, Pratul~P Srinivasan, Rodrigo Ortiz-Cayon, Nima~Khademi
  Kalantari, Ravi Ramamoorthi, Ren Ng, and Abhishek Kar.
\newblock Local light field fusion: Practical view synthesis with prescriptive
  sampling guidelines.
\newblock \emph{ACM Transactions on Graphics (ToG)}, 38\penalty0 (4):\penalty0
  1--14, 2019.

\bibitem[Mildenhall et~al.(2020)Mildenhall, Srinivasan, Tancik, Barron,
  Ramamoorthi, and Ng]{mildenhall2020nerf}
Ben Mildenhall, Pratul~P Srinivasan, Matthew Tancik, Jonathan~T Barron, Ravi
  Ramamoorthi, and Ren Ng.
\newblock Nerf: Representing scenes as neural radiance fields for view
  synthesis.
\newblock In \emph{Computer Vision--ECCV 2020: 16th European Conference,
  Glasgow, UK, August 23--28, 2020, Proceedings, Part I}, pages 405--421, 2020.

\bibitem[M{\"u}ller et~al.(2022)M{\"u}ller, Evans, Schied, and
  Keller]{muller2022instant}
Thomas M{\"u}ller, Alex Evans, Christoph Schied, and Alexander Keller.
\newblock Instant neural graphics primitives with a multiresolution hash
  encoding.
\newblock \emph{ACM Transactions on Graphics (ToG)}, 41\penalty0 (4):\penalty0
  1--15, 2022.

\bibitem[Niemeyer et~al.(2022)Niemeyer, Barron, Mildenhall, Sajjadi, Geiger,
  and Radwan]{niemeyer2022regnerf}
Michael Niemeyer, Jonathan~T Barron, Ben Mildenhall, Mehdi~SM Sajjadi, Andreas
  Geiger, and Noha Radwan.
\newblock Regnerf: Regularizing neural radiance fields for view synthesis from
  sparse inputs.
\newblock In \emph{Proceedings of the IEEE/CVF Conference on Computer Vision
  and Pattern Recognition}, pages 5480--5490, 2022.

\bibitem[Park et~al.(2021{\natexlab{a}})Park, Sinha, Barron, Bouaziz, Goldman,
  Seitz, and Martin-Brualla]{park2021nerfies}
Keunhong Park, Utkarsh Sinha, Jonathan~T Barron, Sofien Bouaziz, Dan~B Goldman,
  Steven~M Seitz, and Ricardo Martin-Brualla.
\newblock Nerfies: Deformable neural radiance fields.
\newblock In \emph{Proceedings of the IEEE/CVF International Conference on
  Computer Vision}, pages 5865--5874, 2021{\natexlab{a}}.

\bibitem[Park et~al.(2021{\natexlab{b}})Park, Sinha, Hedman, Barron, Bouaziz,
  Goldman, Martin-Brualla, and Seitz]{park2021hypernerf}
Keunhong Park, Utkarsh Sinha, Peter Hedman, Jonathan~T Barron, Sofien Bouaziz,
  Dan~B Goldman, Ricardo Martin-Brualla, and Steven~M Seitz.
\newblock Hypernerf: A higher-dimensional representation for topologically
  varying neural radiance fields.
\newblock \emph{arXiv preprint arXiv:2106.13228}, 2021{\natexlab{b}}.

\bibitem[Peng and Chellappa(2023)]{peng2023pdrf}
Cheng Peng and Rama Chellappa.
\newblock Pdrf: progressively deblurring radiance field for fast scene
  reconstruction from blurry images.
\newblock In \emph{Proceedings of the AAAI Conference on Artificial
  Intelligence}, pages 2029--2037, 2023.

\bibitem[Peng et~al.(2024)Peng, Tang, Zhou, Wang, Liu, Li, and
  Chellappa]{peng2024bags}
Cheng Peng, Yutao Tang, Yifan Zhou, Nengyu Wang, Xijun Liu, Deming Li, and Rama
  Chellappa.
\newblock Bags: Blur agnostic gaussian splatting through multi-scale kernel
  modeling.
\newblock \emph{arXiv preprint arXiv:2403.04926}, 2024.

\bibitem[Peng et~al.(2021)Peng, Dong, Wang, Zhang, Shuai, Zhou, and
  Bao]{peng2021animatable}
Sida Peng, Junting Dong, Qianqian Wang, Shangzhan Zhang, Qing Shuai, Xiaowei
  Zhou, and Hujun Bao.
\newblock Animatable neural radiance fields for modeling dynamic human bodies.
\newblock In \emph{Proceedings of the IEEE/CVF International Conference on
  Computer Vision}, pages 14314--14323, 2021.

\bibitem[Pumarola et~al.(2021)Pumarola, Corona, Pons-Moll, and
  Moreno-Noguer]{pumarola2021dnerf}
Albert Pumarola, Enric Corona, Gerard Pons-Moll, and Francesc Moreno-Noguer.
\newblock D-nerf: Neural radiance fields for dynamic scenes.
\newblock In \emph{Proceedings of the IEEE/CVF Conference on Computer Vision
  and Pattern Recognition}, pages 10318--10327, 2021.

\bibitem[Sch{\"o}nberger et~al.(2016)Sch{\"o}nberger, Zheng, Frahm, and
  Pollefeys]{schonberger2016pixelwise}
Johannes~L Sch{\"o}nberger, Enliang Zheng, Jan-Michael Frahm, and Marc
  Pollefeys.
\newblock Pixelwise view selection for unstructured multi-view stereo.
\newblock In \emph{European conference on computer vision}, pages 501--518.
  Springer, 2016.

\bibitem[Shan et~al.(2008)Shan, Jia, and
  Agarwala]{shan2008highmotiondeblurring}
Qi Shan, Jiaya Jia, and Aseem Agarwala.
\newblock High-quality motion deblurring from a single image.
\newblock \emph{Acm transactions on graphics (tog)}, 27\penalty0 (3):\penalty0
  1--10, 2008.

\bibitem[Srinivasan et~al.(2017)Srinivasan, Ng, and
  Ramamoorthi]{srinivasan2017light}
Pratul~P Srinivasan, Ren Ng, and Ravi Ramamoorthi.
\newblock Light field blind motion deblurring.
\newblock In \emph{Proceedings of the IEEE Conference on Computer Vision and
  Pattern Recognition}, pages 3958--3966, 2017.

\bibitem[Sun et~al.(2021)Sun, Xie, Chen, Zhou, and Bao]{sun2021neuralrecon}
Jiaming Sun, Yiming Xie, Linghao Chen, Xiaowei Zhou, and Hujun Bao.
\newblock Neuralrecon: Real-time coherent 3d reconstruction from monocular
  video.
\newblock In \emph{Proceedings of the IEEE/CVF Conference on Computer Vision
  and Pattern Recognition}, pages 15598--15607, 2021.

\bibitem[Tretschk et~al.(2021)Tretschk, Tewari, Golyanik, Zollh{\"o}fer,
  Lassner, and Theobalt]{tretschk2021nonrigid}
Edgar Tretschk, Ayush Tewari, Vladislav Golyanik, Michael Zollh{\"o}fer,
  Christoph Lassner, and Christian Theobalt.
\newblock Non-rigid neural radiance fields: Reconstruction and novel view
  synthesis of a dynamic scene from monocular video.
\newblock In \emph{Proceedings of the IEEE/CVF International Conference on
  Computer Vision}, pages 12959--12970, 2021.

\bibitem[Wang et~al.(2023{\natexlab{a}})Wang, Chen, Loy, and
  Liu]{wang2023sparsenerf}
Guangcong Wang, Zhaoxi Chen, Chen~Change Loy, and Ziwei Liu.
\newblock Sparsenerf: Distilling depth ranking for few-shot novel view
  synthesis.
\newblock In \emph{Proceedings of the IEEE/CVF International Conference on
  Computer Vision}, pages 9065--9076, 2023{\natexlab{a}}.

\bibitem[Wang et~al.(2022)Wang, Wang, Long, Theobalt, Komura, Liu, and
  Wang]{wang2022neuris}
Jiepeng Wang, Peng Wang, Xiaoxiao Long, Christian Theobalt, Taku Komura,
  Lingjie Liu, and Wenping Wang.
\newblock Neuris: Neural reconstruction of indoor scenes using normal priors.
\newblock In \emph{European Conference on Computer Vision}, pages 139--155.
  Springer, 2022.

\bibitem[Wang et~al.(2021)Wang, Liu, Liu, Theobalt, Komura, and
  Wang]{wang2021neus}
Peng Wang, Lingjie Liu, Yuan Liu, Christian Theobalt, Taku Komura, and Wenping
  Wang.
\newblock Neus: Learning neural implicit surfaces by volume rendering for
  multi-view reconstruction.
\newblock \emph{arXiv preprint arXiv:2106.10689}, 2021.

\bibitem[Wang et~al.(2023{\natexlab{b}})Wang, Zhao, Ma, and Liu]{wang2023bad}
Peng Wang, Lingzhe Zhao, Ruijie Ma, and Peidong Liu.
\newblock Bad-nerf: Bundle adjusted deblur neural radiance fields.
\newblock In \emph{Proceedings of the IEEE/CVF Conference on Computer Vision
  and Pattern Recognition}, pages 4170--4179, 2023{\natexlab{b}}.

\bibitem[Wang et~al.(2004)Wang, Bovik, Sheikh, and Simoncelli]{wang2004image}
Zhou Wang, Alan~C Bovik, Hamid~R Sheikh, and Eero~P Simoncelli.
\newblock Image quality assessment: from error visibility to structural
  similarity.
\newblock \emph{IEEE transactions on image processing}, 13\penalty0
  (4):\penalty0 600--612, 2004.

\bibitem[Weng et~al.(2022)Weng, Curless, Srinivasan, Barron, and
  Kemelmacher-Shlizerman]{weng2022humannerf}
Chung-Yi Weng, Brian Curless, Pratul~P Srinivasan, Jonathan~T Barron, and Ira
  Kemelmacher-Shlizerman.
\newblock Humannerf: Free-viewpoint rendering of moving people from monocular
  video.
\newblock In \emph{Proceedings of the IEEE/CVF conference on computer vision
  and pattern Recognition}, pages 16210--16220, 2022.

\bibitem[Whyte et~al.(2012)Whyte, Sivic, Zisserman, and Ponce]{whyte2012non}
Oliver Whyte, Josef Sivic, Andrew Zisserman, and Jean Ponce.
\newblock Non-uniform deblurring for shaken images.
\newblock \emph{International journal of computer vision}, 98:\penalty0
  168--186, 2012.

\bibitem[Wu et~al.(2022)Wu, Li, Peng, Lu, Cao, and Zhong]{wu2022dof}
Zijin Wu, Xingyi Li, Juewen Peng, Hao Lu, Zhiguo Cao, and Weicai Zhong.
\newblock Dof-nerf: Depth-of-field meets neural radiance fields.
\newblock In \emph{Proceedings of the 30th ACM International Conference on
  Multimedia}, pages 1718--1729, 2022.

\bibitem[Wynn and Turmukhambetov(2023)]{wynn2023diffusionerf}
Jamie Wynn and Daniyar Turmukhambetov.
\newblock Diffusionerf: Regularizing neural radiance fields with denoising
  diffusion models.
\newblock In \emph{Proceedings of the IEEE/CVF Conference on Computer Vision
  and Pattern Recognition}, pages 4180--4189, 2023.

\bibitem[Yang et~al.(2023)Yang, Pavone, and Wang]{yang2023freenerf}
Jiawei Yang, Marco Pavone, and Yue Wang.
\newblock Freenerf: Improving few-shot neural rendering with free frequency
  regularization.
\newblock In \emph{Proceedings of the IEEE/CVF Conference on Computer Vision
  and Pattern Recognition}, pages 8254--8263, 2023.

\bibitem[Yifan et~al.(2019)Yifan, Serena, Wu, {\"O}ztireli, and
  Sorkine-Hornung]{yifan2019differentiable}
Wang Yifan, Felice Serena, Shihao Wu, Cengiz {\"O}ztireli, and Olga
  Sorkine-Hornung.
\newblock Differentiable surface splatting for point-based geometry processing.
\newblock \emph{ACM Transactions on Graphics (TOG)}, 38\penalty0 (6):\penalty0
  1--14, 2019.

\bibitem[Yu et~al.(2024)Yu, Chen, Huang, Sattler, and Geiger]{yu2024mip}
Zehao Yu, Anpei Chen, Binbin Huang, Torsten Sattler, and Andreas Geiger.
\newblock Mip-splatting: Alias-free 3d gaussian splatting.
\newblock In \emph{Proceedings of the IEEE/CVF Conference on Computer Vision
  and Pattern Recognition}, pages 19447--19456, 2024.

\bibitem[Zhang et~al.(2018)Zhang, Isola, Efros, Shechtman, and
  Wang]{zhang2018lpips}
Richard Zhang, Phillip Isola, Alexei~A Efros, Eli Shechtman, and Oliver Wang.
\newblock The unreasonable effectiveness of deep features as a perceptual
  metric.
\newblock In \emph{Proceedings of the IEEE conference on computer vision and
  pattern recognition}, pages 586--595, 2018.

\bibitem[Zhao et~al.(2024)Zhao, Wang, and Liu]{zhao2024bad}
Lingzhe Zhao, Peng Wang, and Peidong Liu.
\newblock Bad-gaussians: Bundle adjusted deblur gaussian splatting.
\newblock \emph{arXiv preprint arXiv:2403.11831}, 2024.

\end{thebibliography}
}
\clearpage
\begin{center}
	\textbf{\Large{Appendix}}
	\vspace{5mm}
\end{center}

\section{Implementation Details}
CoCoGaussian is built upon 3DGS~\cite{kerbl20233d} and Deblurring 3DGS~\cite{lee2024deblurring}, trained with a total of 30k iterations with the number of CoC Gaussians, $M$, set to 5. For coarse geometry in the early training stages, $h_{\theta}$ is not trained during the first 2k iterations and begins training afterward. We set $\delta\mathbf{q}_{\max}$ and $\delta\mathbf{s}_{\max}$ to $1.1$. Additionally, after $h_{\theta}$ has been coarsely trained for 4k iterations, CNN $\mathcal{F}$ starts training. Note that, prior to the training of $\mathcal{F}$, the $(M+1)$ output images are averaged to obtain a blurry image. Additionally, we apply a positional encoding layer~\cite{mildenhall2020nerf}, $\gamma$, to 3D points (\textit{i.e.,} $\mathbf{x}_{cam}$ and $\mu_{B}$):
\begin{equation}
	\gamma(\mathbf{x}_{cam})=\left(\sin(2^{k}\pi\mathbf{x}_{cam}),\cos(2^{k}\pi\mathbf{x}_{cam})\right)^{L-1}_{k=0},
\end{equation}
\begin{equation}
	\gamma(\mu_{B})=\left(\sin(2^{k}\pi\mu_{B}),\cos(2^{k}\pi\mu_{B})\right)^{L-1}_{k=0},
\end{equation}
where $L$ denotes the number of the frequencies. The $h_{\theta}$ consists of 3 serial MLP layers, each with 64 hidden units, and parallelized 4 head layers for $K$, $\beta$, $\mathbf{d}$, and $\delta(\mathbf{q}, \mathbf{s})$. The CNN $\mathcal{F}$ comprises 4 convolutional layers with 64 channels each. To compensate for the sparse initial point cloud, we adopt the approach from Deblurring 3DGS, adding approximately 200k additional points after 2.5k iterations and pruning Gaussians based on depth. All experiments are conducted on either an NVIDIA RTX 3090 or NVIDIA V100 GPU.

\begin{figure*}[t]
	\centering
	\includegraphics[width=\linewidth]{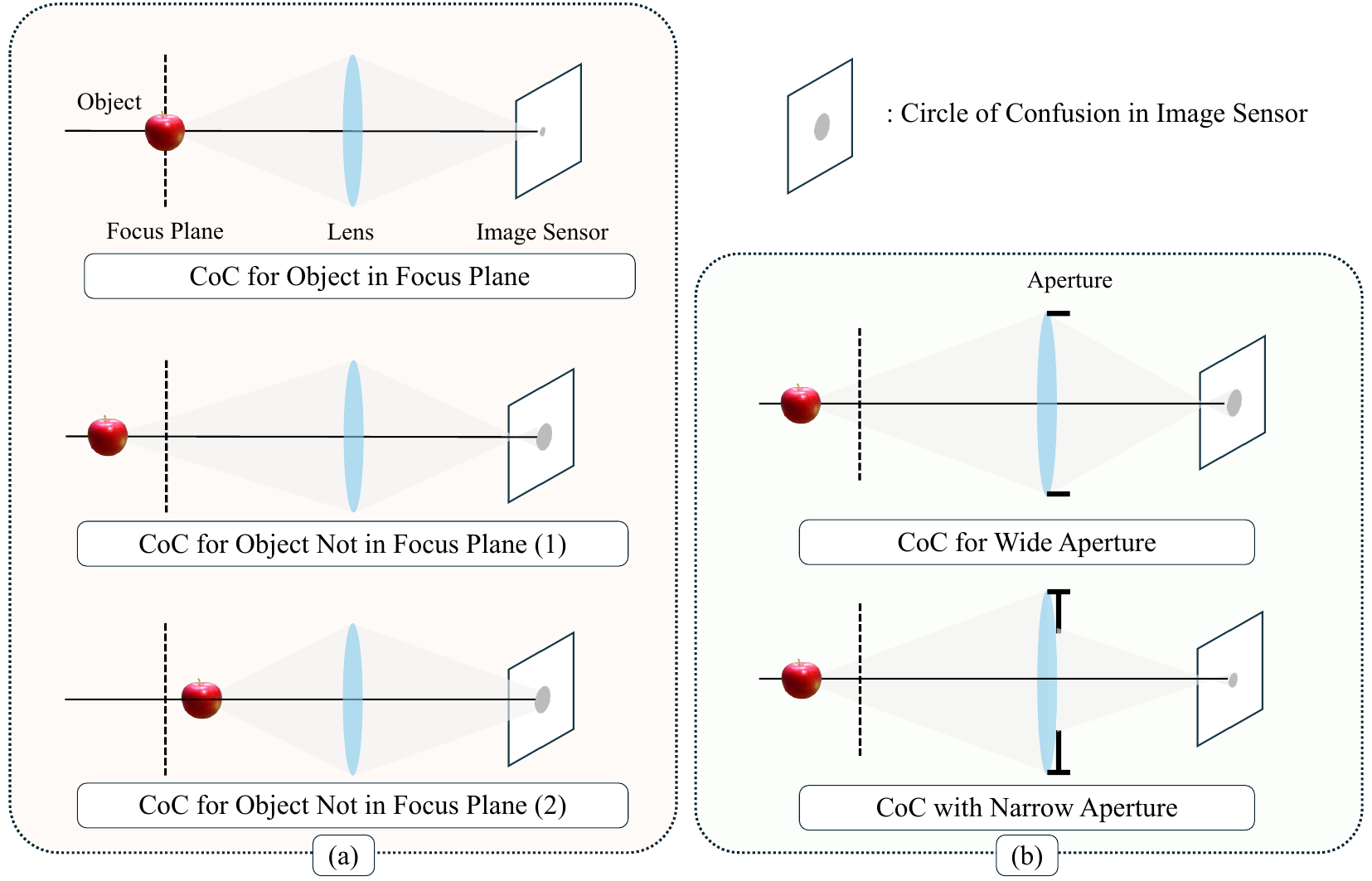}
	\caption{\textbf{(a)} CoC sizes based on the position of object relative to the focus plane, and \textbf{(b)} CoC sizes based on the aperture size.}
	\label{fig:coc_principle}
\end{figure*}

\begin{figure*}[t]
	\centering
	\includegraphics[width=\linewidth]{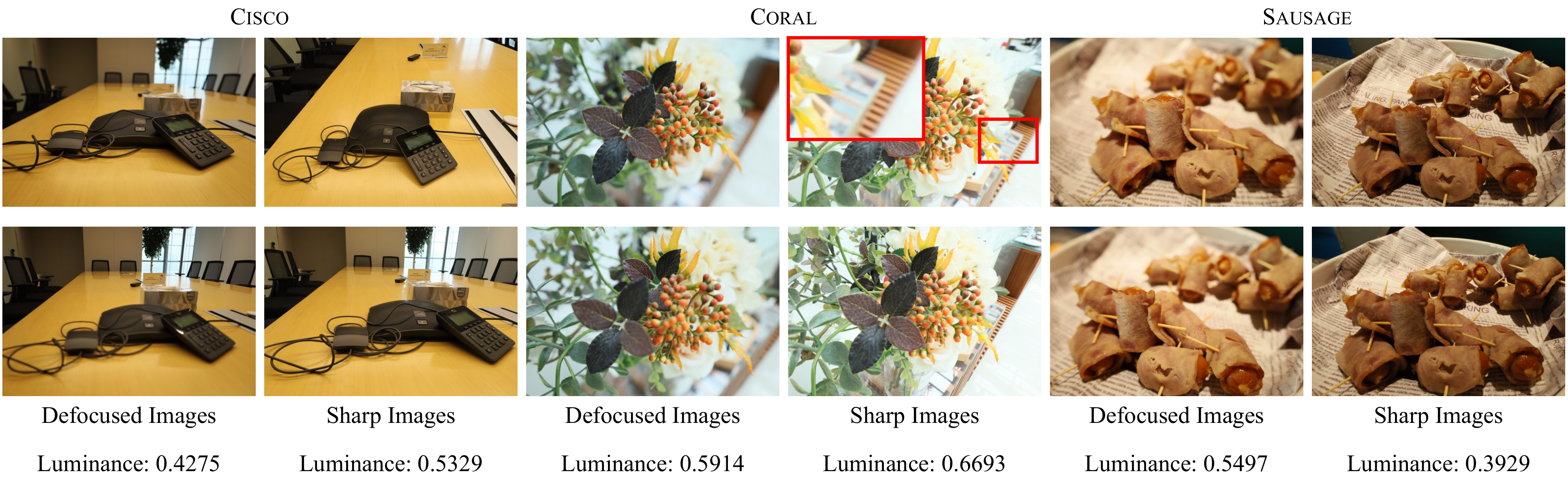}
	\caption{The Luminance Difference between Defocused and Sharp Images. All the luminance values are normalized ranging between 0 and 1.}
	\label{fig:luminance}
\end{figure*}

\section{Circle of Confusion}

In this section, we explain the principles behind the generation of the Circle of Confusion (CoC) based on the focus plane and aperture size. As shown in \cref{fig:coc_principle} \textbf{(a)}, when a subject is precisely located on the focus plane, the radiance emitted from a point on the subject is projected onto the image sensor as a single point. However, when the subject is positioned away from the focus plane, the radiance passes through the lens and forms a circular point spread function on the image sensor. As this circle increases in size, the defocus effect becomes more pronounced. However, even if the subject is off the focus plane, the circle is perceived as a single point if its radius remains below a certain threshold, known as the ``acceptable CoC.'' Thus, even when a CoC exists on the image sensor, the resulting image is perceived as all-in-focus if the CoC is smaller than this threshold.

Additionally, the size of the aperture is another key factor influencing the size of the CoC. As illustrated in \cref{fig:coc_principle} \textbf{(b)}, radiance emitted from a subject passes through the lens and the aperture. With a larger aperture, the radiance forms a larger CoC on the image sensor. With a smaller aperture, only a portion of the light passing through the lens reaches the sensor, resulting in a smaller CoC. Consequently, a smaller aperture produces CoCs smaller than acceptable CoC, leading to an all-in-focus image. However, a smaller aperture also reduces the amount of light reaching the image sensor during the same exposure time, requiring a longer exposure to collect sufficient light. Therefore, capturing an all-in-focus image necessitates (1) a small aperture size and (2) stability to prevent camera movement during the extended exposure time.

\paragraph{Difference from DoF-NeRF~\cite{wu2022dof}.}DoF-NeRF is the first study to apply a physical CoC to 3D scene representation. As it uses NeRF~\cite{mildenhall2020nerf}, a ray tracing-based method, as its backbone, it has the advantage of directly modeling the CoC that reaches the image sensor for a single ray. Specifically, each ray is represented as a CoC on the image sensor, and the color derived from the ray is divided by area of the CoC, ensuring uniform pixel color within a single CoC. However, this approach has three major limitations: (1) it assumes a uniform point spread function (PSF), significantly reducing the flexibility of learning in real-world scenario, (2) it heavily relies on the CoC based on estimated depth, even when derived from uncertain depth, and (3) as an implicit neural representation, its training and rendering are extremely slow.

In contrast, while our CoCoGaussian also models the CoC in different way, it overcomes these three limitations: (1) By generating multiple Gaussians to form the CoC and performing a weighted sum of the resulting images using a CNN $\mathcal{F}$, our approach provides much greater learning flexibility for the PSF. (2) While 3DGS also suffers from challenges with uncertain depth, we propose methods to make CoCoGaussian robust to such depth inaccuracies, as described in \cref{sec:adaptive_gaussian_generation} of the main paper. (3) Since our backbone, 3DGS, is an explicit rasterization-based method, it guarantees fast training and rendering speeds. Thus, while CoCoGaussian draws inspiration from DoF-NeRF, the contributions are clearly distinct and independent.

\section{Deblur-NeRF~\cite{ma2022deblurnerf} Real-World Dataset} \label{sec:deblur_nerf}

As shown in \cref{tab:comparison_deblur} of the main paper, not only our method but also other methods on the Deblur-NeRF~\cite{ma2022deblurnerf} Real-World dataset exhibit relatively poor PSNR and SSIM scores compared to their LPIPS performance. This discrepancy arises from inherent issues within the dataset itself, primarily the luminance differences between the defocused images used for training and the sharp images used for evaluation. As illustrated in \cref{fig:luminance}, the \textsc{Cisco} and \textsc{Coral} scenes have higher luminance in the sharp images, while the \textsc{Sausage} scene has higher luminance in the defocused images. These differences result in lower PSNR and SSIM scores. However, LPIPS evaluates high-level features that align with human visual perception, making it the most reliable metric for this dataset. Consequently, as shown in \cref{tab:comparison_deblur}, CoCoGaussian achieves the best LPIPS score, highlighting the comprehensive performance of our approach.

\section{Additional Ablation Study}
In this section, we conduct two ablative experiments. The first focuses on qualitative results related to the CoC scaling factor $\beta$, and the second evaluates the quantitative results based on the number of CoC Gaussian sets $M$.

\begin{figure}[t]
	\centering
	\includegraphics[width=\linewidth]{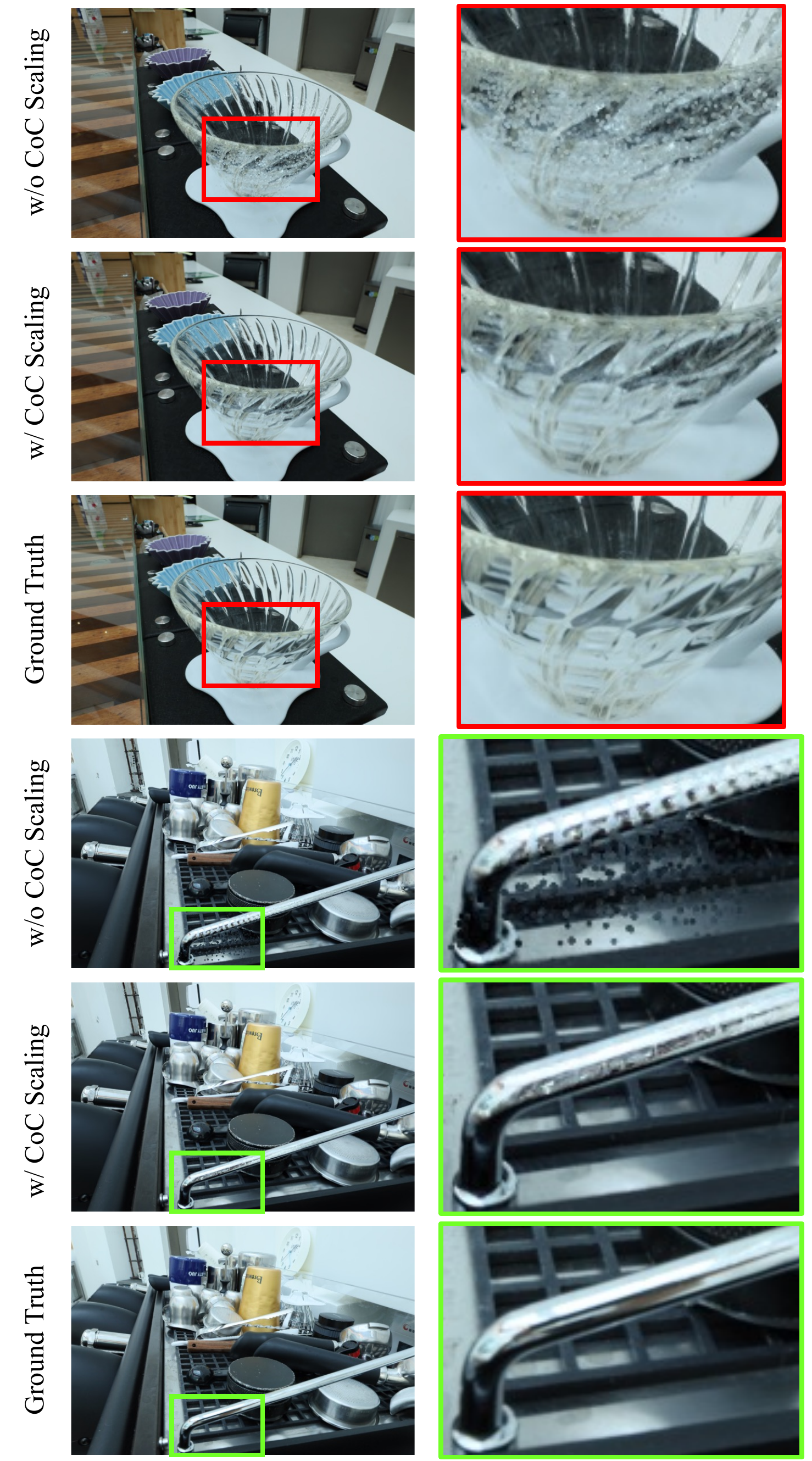}
	\caption{Qualitative Ablation of CoC Scaling Factor. The figures are from \textsc{Cup} and \textsc{Tools} scenes of Deblur-NeRF~\cite{ma2022deblurnerf} real-world dataset.}
	\label{fig:ablation_coc}
\end{figure}

\paragraph{CoC Scaling Factor.}
As shown in \cref{tab:ablation} of our main paper, excluding the CoC scaling factor when modeling CoC Gaussians results in slightly lower performance. This factor is designed to enable robust training of CoC Gaussian positions, even with imperfectly optimized depth, which often occurs in scenes involving reflection or refraction. We visualize scenes with reflections and refractions in \cref{fig:ablation_coc}. The top scene models a transparent cup where light refracts, and the rendered result without CoC scaling shows significant artifacts on the cup. Similarly, in the bottom scene, where a metallic pipe causes light reflection, the absence of CoC scaling leads to many floaters in the affected areas. This indicates that the CoC Gaussians without CoC scaling factor are overfitted to the training images, resulting in poorly optimized base Gaussian positions $\mu_{B}$. Furthermore, this overfitting may potentially affect the covariance of the Gaussians and their SH coefficients. In other words, objects with reflections or refractions often exhibit inconsistent radiance depending on the view direction, making it challenging for 3DGS~\cite{kerbl20233d} to properly optimize for such textures. This inherent limitation of 3DGS causes the CoC Gaussians to overly rely on inaccurately optimized depth, leading to suboptimal outputs. By incorporating the CoC scaling factor during training, we enable robust modeling of CoC Gaussians even in scenes with challenging reflective or refractive surfaces.

\begin{table*}[t]
	\begin{center}
		\caption{Quantitative Results for the Number of CoC Gaussian Sets $M$. The \colorbox{best!25}{orange} and \colorbox{second!35}{yellow} cells respectively indicate the highest and second-highest values.}
		\resizebox{\linewidth}{!}{
			\centering
			\setlength{\tabcolsep}{4pt}
			\renewcommand{\arraystretch}{1.1}
			\scriptsize
			\begin{tabular}{l||c|c|c|c|c|c|c|c|c|c|c|c}
				\toprule 
				
				\multirow{2}{*}{Methods~} 			   	& \multicolumn{3}{c|}{~\textsc{Amiya}~}  	   & \multicolumn{3}{c|}{~\textsc{Book}~} & \multicolumn{3}{c|}{~\textsc{Camera}~} & \multicolumn{3}{c}{~\textsc{Desk}~}  	\\ \cmidrule{2-13}
				&~PSNR$\uparrow$~    &~SSIM$\uparrow$~    &~LPIPS$\downarrow$~ &~PSNR$\uparrow$~    &~SSIM$\uparrow$~    &~LPIPS$\downarrow$~   &~PSNR$\uparrow$~    &~SSIM$\uparrow$~    &~LPIPS$\downarrow$~ &~PSNR$\uparrow$~    &~SSIM$\uparrow$~    &~LPIPS$\downarrow$~  	\\ \midrule
				$M=2$		& 30.06 & 0.9313 & 0.0915 & 30.74 & 0.9277 & 0.0626 & 29.44 & 0.9276 & 0.0725 & \cellcolor{second!35}30.07 & 0.9163 & 0.0652  \\ 
				$M=3$		&\cellcolor{second!35}31.02 & 0.9389 & 0.0882 & \cellcolor{second!35}31.89 & \cellcolor{second!35}0.9377 & \cellcolor{second!35}0.0576 & 29.36 & 0.9270 & 0.0723 & \cellcolor{best!25}30.57 & \cellcolor{best!25}0.9273 & 0.0592	\\ 
				$M=4$		&30.71 & 0.9395 & \cellcolor{second!35}0.0858 & 30.45 & 0.9207 & 0.0600 & \cellcolor{second!35}29.63 & \cellcolor{second!35}0.9283 & 0.0707 & 29.68 & \cellcolor{second!35}0.9193 & \cellcolor{second!35}0.0532	\\ 
				$M=5$		&30.59 & \cellcolor{second!35}0.9406 & 0.0867 & \cellcolor{best!25}31.58 & \cellcolor{best!25}0.9304 & 0.0585 & \cellcolor{best!25}30.21 & \cellcolor{best!25}0.9372 & \cellcolor{best!25}0.0609 & 29.94 & 0.9188 & 0.0559	\\ 
				$M=6$		& \cellcolor{best!25}31.21 & \cellcolor{best!25}0.9411 & \cellcolor{best!25}0.0829 & 30.94 & 0.9255 & \cellcolor{best!25}0.0573 & 29.39 & 0.9266 & \cellcolor{second!35}0.0698 & 29.48 & 0.9151 & \cellcolor{best!25}0.0524	\\ \midrule
				
				\multirow{2}{*}{Methods} 			   	& \multicolumn{3}{c|}{~\textsc{Kendo}~}  	   & \multicolumn{3}{c|}{~\textsc{Plant}~} & \multicolumn{3}{c|}{~\textsc{Shelf}~} & \multicolumn{3}{c}{~\textbf{\textsc{Average}}~}  	\\ \cmidrule{2-13}
				&~PSNR$\uparrow$~    &~SSIM$\uparrow$~    &~LPIPS$\downarrow$~ &~PSNR$\uparrow$~    &~SSIM$\uparrow$~    &~LPIPS$\downarrow$~   &~PSNR$\uparrow$~    &~SSIM$\uparrow$~    &~LPIPS$\downarrow$~ &~PSNR$\uparrow$~    &~SSIM$\uparrow$~    &~LPIPS$\downarrow$~  	\\ \midrule
				$M=2$		& 25.77 & 0.8414 & 0.1303 & \cellcolor{best!25}31.21 & \cellcolor{best!25}0.8899 & 0.0901 & \cellcolor{best!25}32.02 & \cellcolor{best!25}0.9351 & \cellcolor{second!35}0.0455 & 29.90 & \cellcolor{second!35}0.9099 & 0.0797	\\ 
				$M=3$		& 25.83 & 0.8441 & 0.1264 & 30.78 & 0.8814 & 0.0859 & 29.91 & 0.9131 & 0.0462 & \cellcolor{second!35}29.91 & \cellcolor{second!35}0.9099 & 0.0765	\\ 
				$M=4$		& 25.76 & 0.8246 & 0.1277 & 29.59 & 0.8642 & \cellcolor{second!35}0.0830 & 31.20 & \cellcolor{second!35}0.9245 & 0.0480 & 29.57 & 0.9030 & 0.0755	\\ 
				$M=5$		& \cellcolor{second!35}26.26 & \cellcolor{second!35}0.8560 & \cellcolor{second!35}0.1160 & \cellcolor{second!35}31.01 & \cellcolor{second!35}0.8881 & \cellcolor{best!25}0.0693 & 31.38 & 0.9181 & \cellcolor{best!25}0.0431 & \cellcolor{best!25}30.14 & \cellcolor{best!25}0.9127 & \cellcolor{best!25}0.0701	\\ 
				$M=6$		& \cellcolor{best!25}26.67 & \cellcolor{best!25}0.8607 & \cellcolor{best!25}0.1117 & 27.58 & 0.8661 & 0.0894 & \cellcolor{second!35}31.42 & 0.9243 & 0.0460 & 29.53 & 0.9085 & \cellcolor{second!35}0.0728	\\ \bottomrule
			\end{tabular}
		}
		\label{tab:ablation_m}
	\end{center}
	\vspace{-3mm}
\end{table*}

\paragraph{Number of CoC Gaussians.}\label{sec:num_coc}

We conduct ablative experiments on the number of CoC Gaussians, $M$, with the results as shown in \cref{tab:ablation_m}. For PSNR and SSIM, the scores vary inconsistently as $M$ changes. This inconsistency arises primarily from differences in light exposure between defocused images for training and sharp images for evaluation, which depends on the aperture size and exposure time used to equalize the amount of light. In the DoF-NeRF~\cite{wu2022dof} dataset, defocused images are captured with an aperture of f/4 and an exposure time of 1/13 seconds, while sharp images are captured with an aperture of f/11 and an exposure time of 0.8 seconds. The change from f/4 to f/11 reduces light by a factor of 1/8 due to a 3-stop aperture decrease. Meanwhile, the exposure time increases by a factor of approximately 10.4, from 1/13 seconds to 0.8 seconds. Accounting for both aperture and exposure time, sharp images receive 1.3 times more light than defocused images, making the latter slightly darker. Consequently, higher PSNR and SSIM scores do not necessarily indicate better quality.

On the other hand, for LPIPS, larger values of $M$ generally result in better performance. Since LPIPS evaluates quality based on high-level features, such as geometric differences, rather than pixel-level or luminance-based differences, it aligns more closely with human visual perception. As a result, a lower LPIPS score is a more reliable measure of image quality compared to higher PSNR or SSIM scores. Further details are provided in \cref{sec:deblur_nerf}.

\section{CoC Visualization}

\begin{figure*}[t]
	\centering
	\includegraphics[width=\linewidth]{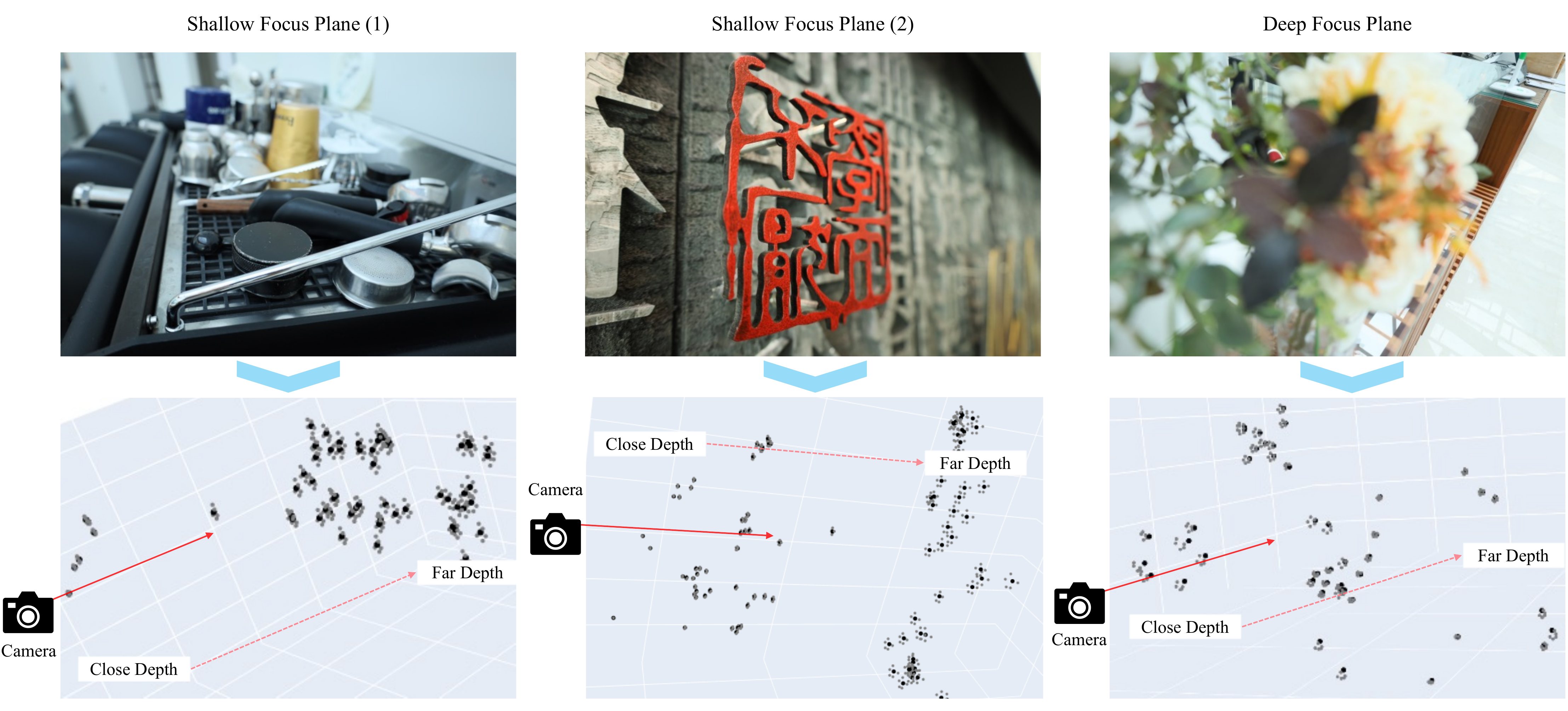}
	\caption{Gaussian Positions of Defocused Images. The black and gray dots indicate base and CoC Gaussians, respectively.}
	\label{fig:coc_visualize}
	\vspace{-30mm}
\end{figure*}

\begin{figure*}[h]
	\centering
	\includegraphics[width=0.9\columnwidth]{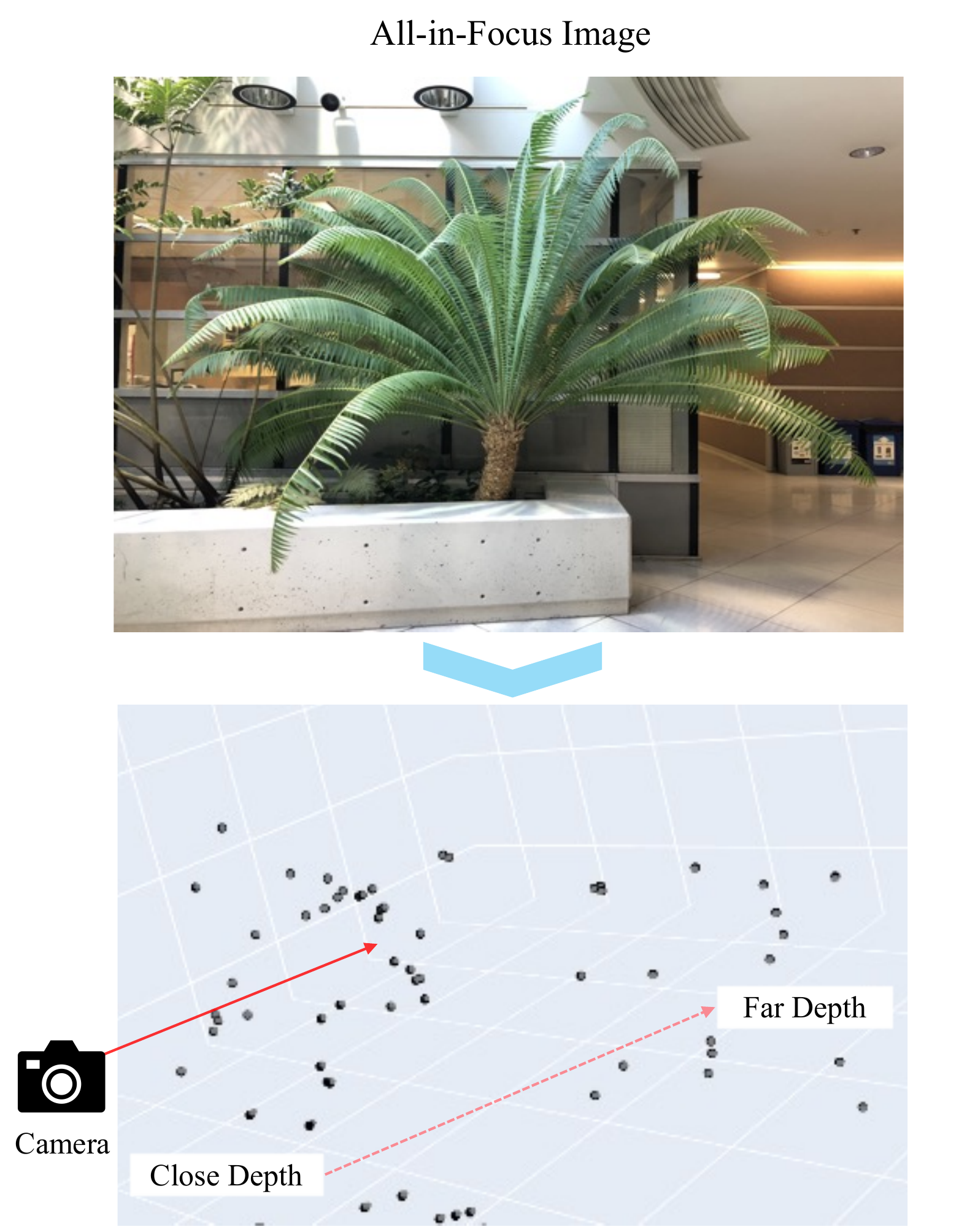}
	\caption{Gaussian Positions of All-in-Focus Images.}
	\label{fig:coc_visualize_nerf}
\end{figure*}

We visualize the CoC for various types of images in \cref{fig:coc_visualize} and \cref{fig:coc_visualize_nerf}. To simplify the visualization, we randomly sample a subset of positions from numerous Gaussians. The points in \cref{fig:coc_visualize} represent the positions of Gaussians for defocused images. For images where the focus plane is close to the camera, CoCoGaussian generates CoC sizes that are very small for shallow depths and progressively larger for deeper depths. However, when the focus plane is farther from the camera, the CoC sizes reconstructed at greater depths are smaller. This demonstrates the effectiveness of our modeling, accurately reflecting the principles of defocus blur.

In contrast, \cref{fig:coc_visualize_nerf} visualizes the CoC for an all-in-focus scene~\cite{mildenhall2020nerf}, where the CoC sizes remain uniformly small regardless of depth. This suggests that the learned aperture size is very small, ensuring precise modeling of all-in-focus images. In other words, CoCoGaussian can model small apertures and capture subtle defocus effects that are imperceptible to the human eye, which contributes to its superior performance compared to naive 3DGS~\cite{kerbl20233d}, as demonstrated in \cref{tab:ablation_m} of the main paper. In summary, our model can accurately represent 3D scenes for both defocused and all-in-focus images, highlighting its versatility and robustness.

\begin{table}[t] 
	\begin{center}
		\caption{Computational Efficiency and Speed. * indicates that the rendering speed is identical to that of the corresponding model.}
		\vspace{-2mm}
		\resizebox{\columnwidth}{!}{
			\centering
			\setlength{\tabcolsep}{5pt}
			\begin{tabular}{l||c|c|c}
				\toprule
				\multirow{2}{*}{Methods} 		& ~GPU Memory~	   	& ~Training Time~ & \multirow{2}{*}{~Rendering Speed~}  	   \\ 
				& (GB) & (min) &  \\ \midrule
				BAGS~\cite{peng2024bags}	& 4.7& 47  & *Mip-Splatting~\cite{yu2024mip}\\ 
				CoCoGaussian~	& 5.3& 45 &*3DGS~\cite{kerbl20233d} \\ \bottomrule
			\end{tabular}
		}
		\label{tab:speed}
	\end{center}
	\vspace{-7mm}
\end{table}

\section{Computational Efficiency and Speed}
We compare our GPU usage, training time, and rendering speed with BAGS~\cite{peng2024bags}, a state-of-the-art method, on the Deblur-NeRF real-world dataset using an NVIDIA RTX 3090. As shown in the \cref{tab:speed}, CoCoGaussian achieves comparable resource consumption and training time while delivering superior performance.

After the training phase, CoCoGaussian renders sharp images using only the $\mathbf{G}_{B}$ through a naive 3DGS. In the other words, the rendering speed and memory cost are the same as 3DGS alone. Therefore, our method is feasible for real-time applications that rely on 3DGS rendering speeds.

\clearpage


\clearpage

\section{Per-Scene Quantitative Results}

We present the performance for individual scenes across all datasets in \cref{tab:per_scene_deblur_nerf_synthetic,tab:per_scene_deblur_nerf_real,tab:per_scene_dof_nerf_real}. CoCoGaussian achieves the best LPIPS scores in all scenes except for the \textsc{Coral} scene in the Deblur-NeRF Real-World dataset. As discussed in \cref{sec:num_coc,sec:deblur_nerf}, this indicates that CoCoGaussian exhibits the most superior high-level feature representation.

\begin{table*}[!t] 
	\begin{center}
		\caption{Per-Scene Quantitative Results on Deblur-NeRF~\cite{ma2022deblurnerf} Synthetic Dataset.}
		\resizebox{\linewidth}{!}{
			\centering
			\setlength{\tabcolsep}{1.2pt}
			\renewcommand{\arraystretch}{1.1}
			\scriptsize
			\begin{tabular}{l||c|c|c|c|c|c|c|c|c|c|c|c|c|c|c}
				\toprule 
				
				\multirow{2}{*}{Methods~} 			   	& \multicolumn{3}{c|}{~\textsc{Factory}~}  	   & \multicolumn{3}{c|}{~\textsc{Cozyroom}~} & \multicolumn{3}{c|}{~\textsc{Pool}~} & \multicolumn{3}{c|}{~\textsc{Tanabata}~} & \multicolumn{3}{c}{~\textsc{Trolley}~}  	\\ \cmidrule{2-16}
				&~PSNR$\uparrow$~    &~SSIM$\uparrow$~    &~LPIPS$\downarrow$~ &~PSNR$\uparrow$~    &~SSIM$\uparrow$~    &~LPIPS$\downarrow$~   &~PSNR$\uparrow$~    &~SSIM$\uparrow$~    &~LPIPS$\downarrow$~ &~PSNR$\uparrow$~    &~SSIM$\uparrow$~    &~LPIPS$\downarrow$~  	&~PSNR$\uparrow$~    &~SSIM$\uparrow$~    &~LPIPS$\downarrow$~ \\ \midrule
				NeRF~\cite{mildenhall2020nerf}		&25.36 & 0.7847 & 0.2351 & 30.03 & 0.8926 & 0.0885 & 27.77 & 0.7266 & 0.3340 & 23.90 & 0.7811 & 0.2142 & 22.67 & 0.7103 & 0.2799	\\ 
				3DGS~\cite{kerbl20233d}		& 24.52 & 0.8057 & 0.1842 & 30.09 & 0.9024 & 0.0692 & 20.14 & 0.4451 & 0.5094 & 23.08 & 0.7981 & 0.1710 & 22.26 & 0.7400 & 0.2281 \\ \midrule
				Deblur-NeRF~\cite{ma2022deblurnerf}		&28.03 & 0.8628 & 0.1127 & 31.85 & 0.9175 & 0.0481 & 30.52 & 0.8246 & 0.1901 & 26.26 & 0.8517 & 0.0995 & 25.18 & 0.8067 & 0.1436	\\ 
				PDRF-10~\cite{peng2023pdrf}		&\cellcolor{best!25}30.90 & 0.9138 & 0.1066 & 32.29 & 0.9305 & 0.0518 & 30.97 & 0.8408 & 0.1893 & 28.18 & 0.9006 & 0.0819 & 28.07 & 0.8799 & 0.1210	\\ 
				DP-NeRF~\cite{lee2023dp}		& 29.26 & 0.8793 & 0.1035 & 32.11 & 0.9215 & 0.0386 & 31.44 & 0.8529 & 0.1563 & 27.05 & 0.8635 & 0.0779 & 26.79 & 0.8395 & 0.1170	\\ \midrule
				Deblurring 3DGS~\cite{lee2024deblurring}		& 27.39 & 0.8922 & 0.1160 & 31.29 & 0.9201 & 0.0505 & 31.27 & 0.8565 & 0.1556 & 27.04 & 0.9029 & 0.0872 & 27.53 & 0.8843 & 0.1167	\\
				BAGS~\cite{peng2024bags}		& \cellcolor{second!35}30.87 & \cellcolor{best!25}0.9334 & 0.0724 & \cellcolor{second!35}32.45 & \cellcolor{second!35}0.9312 & \cellcolor{second!35}0.0289 & \cellcolor{second!35}31.78 & \cellcolor{second!35}0.8645 & \cellcolor{second!35}0.0932 & \cellcolor{second!35}29.19 & \cellcolor{second!35}0.9278 & \cellcolor{second!35}0.0405 & \cellcolor{second!35}28.97 & \cellcolor{second!35}0.9070 & \cellcolor{second!35}0.0804	\\ \midrule
				Ours		& 30.15 & \cellcolor{second!35}0.9300 & \cellcolor{best!25}0.0489 & \cellcolor{best!25}33.02 & \cellcolor{best!25}0.9410 & \cellcolor{best!25}0.0213 & \cellcolor{best!25}31.96 & \cellcolor{best!25}0.8788 & \cellcolor{best!25}0.0803 & \cellcolor{best!25}29.65 & \cellcolor{best!25}0.9396 & \cellcolor{best!25}0.0263 & \cellcolor{best!25}29.41 & \cellcolor{best!25}0.9165 & \cellcolor{best!25}0.0620	\\ \bottomrule
			\end{tabular}
		}
		\label{tab:per_scene_deblur_nerf_synthetic}
	\end{center}
	\vspace{-3mm}
\end{table*}

\begin{table*}[!t] 
	\begin{center}
		\caption{Per-Scene Quantitative Results on Deblur-NeRF~\cite{ma2022deblurnerf} Real-World Dataset.}
		\resizebox{\linewidth}{!}{
			\centering
			\setlength{\tabcolsep}{1.2pt}
			\renewcommand{\arraystretch}{1.1}
			\scriptsize
			\begin{tabular}{l||c|c|c|c|c|c|c|c|c|c|c|c|c|c|c}
				\toprule 
				
				\multirow{2}{*}{Methods~} 			   	& \multicolumn{3}{c|}{~\textsc{Cake}~}  	   & \multicolumn{3}{c|}{~\textsc{Caps}~} & \multicolumn{3}{c|}{~\textsc{Cisco}~} & \multicolumn{3}{c|}{~\textsc{Coral}~} & \multicolumn{3}{c}{~\textsc{Cupcake}~}  	\\ \cmidrule{2-16}
				&~PSNR$\uparrow$~    &~SSIM$\uparrow$~    &~LPIPS$\downarrow$~ &~PSNR$\uparrow$~    &~SSIM$\uparrow$~    &~LPIPS$\downarrow$~   &~PSNR$\uparrow$~    &~SSIM$\uparrow$~    &~LPIPS$\downarrow$~ &~PSNR$\uparrow$~    &~SSIM$\uparrow$~    &~LPIPS$\downarrow$~  	&~PSNR$\uparrow$~    &~SSIM$\uparrow$~    &~LPIPS$\downarrow$~ \\ \midrule
				NeRF~\cite{mildenhall2020nerf}		&24.42 & 0.7210 & 0.2250 & 22.73 & 0.6312 & 0.2801 & 20.72 & 0.7217 & 0.1256 & 19.81 & 0.5658 & 0.2155 & 21.88 & 0.6809 & 0.2689	\\ 
				3DGS~\cite{kerbl20233d}		& 20.16 & 0.5903 & 0.2082 & 19.08 & 0.4355 & 0.4329 & 20.01 & 0.6931 & 0.1781 & 19.50 & 0.5519 & 0.3111 & 21.53 & 0.6794 & 0.2081	\\ \midrule
				Deblur-NeRF~\cite{ma2022deblurnerf}		& 26.27 & 0.7800 & 0.1282 & 23.87 & 0.7128 & 0.1612 & \cellcolor{best!25}20.83 & \cellcolor{second!35}0.7270 & 0.0868 & 19.85 & 0.5999 & 0.1160 & 22.26 & 0.7219 & 0.1214	\\ 
				PDRF-10~\cite{peng2023pdrf}		&\cellcolor{best!25}27.06 & 0.8032 & 0.1622 & 24.06 & 0.7102 & 0.2854 & 20.68 & 0.7239 & 0.0943 & 19.61 & 0.5894 & 0.2335 & \cellcolor{best!25}22.95 & \cellcolor{second!35}0.7421 & 0.1862	\\ 
				DP-NeRF~\cite{lee2023dp}		& 26.16 & 0.7781 & 0.1267 & 23.95 & 0.7122 & 0.1430 & \cellcolor{second!35}20.73 & 0.7260 & 0.0840 & \cellcolor{best!25}20.11 & \cellcolor{best!25}0.6107 & \cellcolor{best!25}0.0960 & \cellcolor{second!35}22.80 & 0.7409 & 0.1178	\\ \midrule
				Deblurring 3DGS~\cite{lee2024deblurring}		&\cellcolor{second!35}26.91 & \cellcolor{second!35}0.8039 & 0.1136 & \cellcolor{second!35}24.45 & 0.7391 & 0.1509 & 20.55 & 0.7227 & 0.0816 & 18.99 & 0.5534 & 0.2767 & 22.11 & 0.7356 & \cellcolor{second!35}0.1021	\\
				BAGS~\cite{peng2024bags}		& 26.53 & 0.7996 & \cellcolor{second!35}0.1113 & 24.15 & \cellcolor{second!35}0.7422 & \cellcolor{second!35}0.1391 & 20.31 & 0.7230 & \cellcolor{second!35}0.0759 & \cellcolor{second!35}19.63 & \cellcolor{second!35}0.6016 & 0.1114 & 21.52 & 0.6971 & 0.1214	\\ \midrule
				Ours		&26.65 & \cellcolor{best!25}0.8043 & \cellcolor{best!25}0.1037 & \cellcolor{best!25}24.62 & \cellcolor{best!25}0.7472 & \cellcolor{best!25}0.1334 & \cellcolor{best!25}20.83 & \cellcolor{best!25}0.7359 & \cellcolor{best!25}0.0680 & 19.57 & 0.5925 & \cellcolor{second!35}0.1105 & 22.48 & \cellcolor{best!25}0.7515 & \cellcolor{best!25}0.0739	\\ \midrule \midrule
				
				\multirow{2}{*}{Methods~} 			   	& \multicolumn{3}{c|}{~\textsc{Cups}~}  	   & \multicolumn{3}{c|}{~\textsc{Daisy}~} & \multicolumn{3}{c|}{~\textsc{Sausage}~} & \multicolumn{3}{c|}{~\textsc{Seal}~} & \multicolumn{3}{c}{~\textsc{Tools}~}  	\\ \cmidrule{2-16}
				&~PSNR$\uparrow$~    &~SSIM$\uparrow$~    &~LPIPS$\downarrow$~ &~PSNR$\uparrow$~    &~SSIM$\uparrow$~    &~LPIPS$\downarrow$~   &~PSNR$\uparrow$~    &~SSIM$\uparrow$~    &~LPIPS$\downarrow$~ &~PSNR$\uparrow$~    &~SSIM$\uparrow$~    &~LPIPS$\downarrow$~  	&~PSNR$\uparrow$~    &~SSIM$\uparrow$~    &~LPIPS$\downarrow$~ \\ \midrule
				NeRF~\cite{mildenhall2020nerf}		&25.02 & 0.7581 & 0.2315 & 22.74 & 0.6203 & 0.2621 & 17.79 & 0.4830 & 0.2789 & 22.79 & 0.6267 & 0.2680 & 26.08 & 0.8523 & 0.1547	\\ 
				3DGS~\cite{kerbl20233d}		& 20.55 & 0.6459 & 0.3211 & 20.96 & 0.6004 & 0.2629 & 17.83 & 0.4718 & 0.2855 & 22.25 & 0.5905 & 0.3057 & 23.82 & 0.8050 & 0.1953	\\ \midrule
				Deblur-NeRF~\cite{ma2022deblurnerf}		& 26.21 & 0.7987 & 0.1271 & 23.52 & 0.6870 & 0.1208 & 18.01 & 0.4998 & 0.1796 & 26.04 & 0.7773 & 0.1048 & 27.81 & 0.8949 & 0.0610	\\ 
				PDRF-10~\cite{peng2023pdrf}		& \cellcolor{second!35}26.39 & 0.8066 & 0.1370 & \cellcolor{best!25}24.49 & \cellcolor{best!25}0.7426 & 0.1024 & \cellcolor{second!35}18.94 & 0.5686 & 0.2126 & \cellcolor{best!25}26.36 & 0.7959 & 0.1927 & 28.00 & 0.8995 & 0.1395	\\ 
				DP-NeRF~\cite{lee2023dp}		&\cellcolor{best!25}26.75 & 0.8136 & 0.1035 & \cellcolor{second!35}23.79 & 0.6971 & 0.1075 & 18.35 & 0.5443 & 0.1473 & 25.95 & 0.7779 & 0.1026 & \cellcolor{second!35}28.07 & 0.8980 & 0.0539	\\ \midrule
				Deblurring 3DGS~\cite{lee2024deblurring}		& 26.23 & \cellcolor{second!35}0.8230 & 0.1014 & 23.39 & 0.7288 & 0.0979 & 18.83 & 0.5609 & 0.1470 & 26.04 & \cellcolor{second!35}0.8087 & 0.0988 & 27.86 & 0.9069 & 0.0619	\\
				BAGS~\cite{peng2024bags}		& 26.14 & 0.8194 & \cellcolor{second!35}0.0901 & 23.00 & 0.7332 & \cellcolor{second!35}0.0540 & 18.66 & \cellcolor{second!35}0.5721 & \cellcolor{second!35}0.1176 & \cellcolor{second!35}26.16 & 0.8050 & \cellcolor{second!35}0.0967 & \cellcolor{best!25}28.72 & \cellcolor{second!35}0.9148 & \cellcolor{second!35}0.0450	\\ \midrule
				Ours		&26.20 & \cellcolor{best!25}0.8317 & \cellcolor{best!25}0.0773 & 23.39 & \cellcolor{second!35}0.7425 & \cellcolor{best!25}0.0503 & \cellcolor{best!25}19.20 & \cellcolor{best!25}0.5864 & \cellcolor{best!25}0.1007 & 26.10 & \cellcolor{best!25}0.8243 & \cellcolor{best!25}0.0663 & 27.91 & \cellcolor{best!25}0.9149 & \cellcolor{best!25}0.0416	\\ \bottomrule
			\end{tabular}
		}
		\label{tab:per_scene_deblur_nerf_real}
	\end{center}
	\vspace{-3mm}
\end{table*}

\begin{table*}[!t] 
	\begin{center}
		\caption{Per-Scene Quantitative Results on DoF-NeRF~\cite{wu2022dof} Real-World Dataset.}
		\resizebox{0.9\linewidth}{!}{
			\centering
			\setlength{\tabcolsep}{3pt}
			\renewcommand{\arraystretch}{1.1}
			\scriptsize
			\begin{tabular}{l||c|c|c|c|c|c|c|c|c|c|c|c}
				\toprule 
				
				\multirow{2}{*}{Methods~} 			   	& \multicolumn{3}{c|}{~\textsc{Amiya}~}  	   & \multicolumn{3}{c|}{~\textsc{Book}~} & \multicolumn{3}{c|}{~\textsc{Camera}~} & \multicolumn{3}{c}{~\textsc{Desk}~}  	\\ \cmidrule{2-13}
				&~PSNR$\uparrow$~    &~SSIM$\uparrow$~    &~LPIPS$\downarrow$~ &~PSNR$\uparrow$~    &~SSIM$\uparrow$~    &~LPIPS$\downarrow$~   &~PSNR$\uparrow$~    &~SSIM$\uparrow$~    &~LPIPS$\downarrow$~ &~PSNR$\uparrow$~    &~SSIM$\uparrow$~    &~LPIPS$\downarrow$~  	\\ \midrule
				
				3DGS~\cite{kerbl20233d}		&26.16 & 0.8718 & 0.1761 & 24.73 & 0.8480 & 0.1980 & 25.05 & 0.8429 & 0.1641 & 27.47 & 0.8760 & 0.1422	\\ \midrule
				DP-NeRF~\cite{ma2022deblurnerf}		& 28.64 & 0.8849 & 0.1421 & 28.51 & 0.8520 & 0.1669 & 26.50 & 0.8667 & 0.1332 & 27.57 & 0.8238 & 0.1712	\\ \midrule
				Deblurring 3D-GS~\cite{peng2023pdrf}		& 26.21 & 0.8728 & 0.1815 & 27.29 & \cellcolor{second!35}0.8754 & \cellcolor{second!35}0.1584 & 25.63 & 0.8486 & 0.1796 & 29.08 & 0.8478 & 0.1672	\\ 
				BAGS~\cite{lee2023dp}		&\cellcolor{best!25}31.86 & \cellcolor{best!25}0.9453 & 0.0874 & \cellcolor{second!35}29.41 & 0.7515 & 0.2214 & \cellcolor{second!35}29.20 & \cellcolor{second!35}0.9248 & \cellcolor{second!35}0.0730 & \cellcolor{second!35}29.77 & \cellcolor{second!35}0.9175 & \cellcolor{second!35}0.0722	\\ \midrule
				Ours		& \cellcolor{second!35}30.59 & \cellcolor{second!35}0.9406 & \cellcolor{best!25}0.0867 & \cellcolor{best!25}31.58 & \cellcolor{best!25}0.9304 & \cellcolor{best!25}0.0585 & \cellcolor{best!25}30.21 & \cellcolor{best!25}0.9372 & \cellcolor{best!25}0.0609 & \cellcolor{best!25}29.94 & \cellcolor{best!25}0.9188 & \cellcolor{best!25}0.0559	\\ \midrule \midrule

				\multirow{2}{*}{Methods} 			   	& \multicolumn{3}{c|}{~\textsc{Kendo}~}  	   & \multicolumn{3}{c|}{~\textsc{Plant}~} & \multicolumn{3}{c|}{~\textsc{Shelf}~} & \multicolumn{3}{c}{~\textbf{\textsc{Average}}~}  	\\ \cmidrule{2-13}
				&~PSNR$\uparrow$~    &~SSIM$\uparrow$~    &~LPIPS$\downarrow$~ &~PSNR$\uparrow$~    &~SSIM$\uparrow$~    &~LPIPS$\downarrow$~   &~PSNR$\uparrow$~    &~SSIM$\uparrow$~    &~LPIPS$\downarrow$~ &~PSNR$\uparrow$~    &~SSIM$\uparrow$~    &~LPIPS$\downarrow$~  	\\ \midrule
				3DGS~\cite{kerbl20233d}		&20.63 & 0.6839 & 0.2715 & 26.64 & 0.7897 & 0.2047 & 29.37 & 0.8913 & 0.1150 & 25.72 & 0.8291 & 0.1817	\\ \midrule
				DP-NeRF~\cite{ma2022deblurnerf}		&19.64 & 0.6166 & 0.3205 & 28.07 & 0.8158 & 0.1668 & 28.64 & 0.8415 & 0.1525 & 26.80 & 0.8145 & 0.1790	\\ \midrule
				Deblurring 3D-GS~\cite{peng2023pdrf}		&22.74 & 0.7559 & 0.1904 & 26.66 & 0.7900 & 0.2081 & 28.44 & 0.8798 & 0.1102 & 26.58 & 0.8386 & 0.1708	\\ 
				BAGS~\cite{lee2023dp}		& \cellcolor{second!35}26.19 & \cellcolor{second!35}0.8457 & \cellcolor{second!35}0.1291 & \cellcolor{second!35}30.66 & \cellcolor{second!35}0.8644 & \cellcolor{second!35}0.1128 & \cellcolor{best!25}32.03 & \cellcolor{best!25}0.9223 & \cellcolor{second!35}0.0741 & \cellcolor{second!35}29.87 & \cellcolor{second!35}0.8816 & \cellcolor{second!35}0.1100	\\ \midrule
				Ours		&\cellcolor{best!25}26.26 & \cellcolor{best!25}0.8560 & \cellcolor{best!25}0.1160 & \cellcolor{best!25}31.01 & \cellcolor{best!25}0.8881 & \cellcolor{best!25}0.0693 & \cellcolor{second!35}31.38 & \cellcolor{second!35}0.9181 & \cellcolor{best!25}0.0431 & \cellcolor{best!25}30.14 & \cellcolor{best!25}0.9127 & \cellcolor{best!25}0.0701	\\ \bottomrule

			\end{tabular}
		}
		\label{tab:per_scene_dof_nerf_real}
	\end{center}
	\vspace{-3mm}
\end{table*}



\end{document}